\documentclass[journal,twoside,web]{ieeecolor}
\usepackage{generic}
\usepackage{cite}
\usepackage{amsmath,amssymb,amsfonts}
\usepackage{algorithmic}
\usepackage{graphicx}
\usepackage{algorithm,algorithmic}
\usepackage{hyperref}
\usepackage{textcomp}

\usepackage{framed,multirow}
\usepackage{latexsym}
\usepackage{xcolor}
\usepackage{booktabs}
\usepackage{caption}
\usepackage[font=footnotesize,labelformat=simple]{subcaption}

\let\labelindent\relax

\usepackage{siunitx}

\usepackage{enumitem}

\def\BibTeX{{\rm B\kern-.05em{\sc i\kern-.025em b}\kern-.08em
    T\kern-.1667em\lower.7ex\hbox{E}\kern-.125emX}}
\begin{document}


\title{Deep Learning Techniques for Automatic Lateral X-ray Cephalometric Landmark Detection: Is the Problem Solved?}

\author{Hongyuan Zhang, Ching-Wei Wang, Hikam Muzakky, Juan Dai, Xuguang Li,
Chenglong Ma, Qian Wu, Xianan Cui, Kunlun Xu, Pengfei He, Dongqian Guo, Xianlong Wang, Hyunseok Lee, Zhangnan Zhong, Zhu Zhu
and Bingsheng Huang
\thanks{
This study received support from the National Science and Technology Council, Taiwan (NSTC 113-222-E-011-MY3, 112-2221-E-011-052), Shenzhen-Hong Kong Institute of Brain Science-Shenzhen Fundamental Research Institutions (No. 2023SHIBS0003), Medicine-Engineering Interdisciplinary Research Foundation of ShenZhen University and National Natural Science Foundation of China (No. 62371303).
Special thanks to the author of the MMpose package \cite{contributors2020openmmlab} and the technical support provided by OpenMMLab.
Last but not least,
the authors would also like to thank all the other CL-Detection2023 Challenge participants that registered and submitted their algorithm Docker containers. 
(Hongyuan Zhang, Ching-Wei Wang, Hikam Muzakky and Juan Dai contributed equally to this work as joint first authors) (Corresponding author: Ching-Wei Wang, e-mail: cweiwang@mail.ntust.edu.tw; Bingsheng Huang, e-email: huangb@szu.edu.cn.)
}
\thanks{H. Zhang and B. Huang are with Medical AI Lab, School of Biomedical Engineering, Shenzhen University Medical School, Shenzhen University, Shenzhen, 518060, China and Guangdong Key Laboratory of Biomedical Measurements and Ultrasound Imaging, School of Biomedical Engineering, Shenzhen University Medical School, Shenzhen University, Shenzhen, 518060, China.}
\thanks{C. Wang and H. Muzakky are Graduate Institute of Biomedical Engineering, National Taiwan University of Science and Technology, Taiwan.}
\thanks{J. Dai and X. Li are with Department of Stomatology, Shenzhen University General Hospital, Shenzhen University, Shenzhen, 518055, China and Institute of Stomatological Research, Shenzhen University, Shenzhen, 518055, China.}
\thanks{
C. Ma is with Chohotech Technology Co., Ltd., Lianchuang Street, Yuhang District, Hangzhou, 311422, China.
Q. Wu is with Information Systems Technology and Design (ISTD) Pillar, Singapore University of Technology and Design (SUTD), Singapore.
X. Cui is with School of Computer Science, Wuhan University, Wuhan, 430072, China. 
K. Xu is with Wangxuan Institute of Computer Technology, Peking University, Beijing, 100871,
China.
P. He is with School of Artificial Intelligence, Xidian University, Xi’an, 710126, China.
D. Guo is with State Key Laboratory of Internet of Things for Smart City, University of Macau, Taipa, Macau.
X. Wang is with The Faculty of Information Science and Engineering, Ocean University of China, Qingdao, 266100, China.
H. Lee is with Digital Healthcare Department, Daegu-Gyeongbuk Medical Innovation Foundation, Daegu, 427724, Korea.
Z. Zhong is with School of Life Science, South China Normal University, Guangzhou, 510631, China. 
Z. Zhu is with Information and Data Department, Children’s Hospital Zhejiang University School of Medicine, Hangzhou, 310000, China.}
}

\maketitle

\begin{abstract}
Localization of the craniofacial landmarks from lateral cephalograms is a fundamental task in cephalometric analysis.
The automation of the corresponding tasks has thus been the subject of intense research over the past decades.
In this paper,
we introduce the “\textbf{C}ephalometric \textbf{L}andmark \textbf{Detection} (CL-Detection)” dataset, which is
the largest publicly available and comprehensive dataset for cephalometric landmark detection.
This multi-center and multi-vendor dataset includes 600 lateral X-ray images with 38 landmarks acquired with different equipment from three medical centers.
The overarching objective of this paper is to measure how far state-of-the-art deep learning methods can go for cephalometric landmark detection.
Following the 2023 MICCAI CL-Detection Challenge,
we report the results of the top ten research groups using deep learning methods.
Results show that the best methods closely approximate the expert analysis, 
achieving a mean detection rate of 75.719\% and a mean radial error of 1.518 mm.
While there is room for improvement,
these findings undeniably open the door to highly accurate and fully automatic location of craniofacial landmarks.
We also identify scenarios for which deep learning methods are still failing.
Both the dataset and detailed results are publicly available online, while the platform will remain open for the community to benchmark future algorithm developments at \url{https://cl-detection2023.grand-challenge.org/}.
\end{abstract}


\section{Introduction} \label{sec:introduction}

\IEEEPARstart{C}{ephalometric} analysis is a fundamental examination which is routinely used in fields of orthodontics and orthognathics \cite{proffit2006contemporary,legan1980soft}.
Over the years, various analysis methods have been proposed for cephalometric analysis,
such as Ricketts analysis \cite{ricketts1982orthodontic}, Downs analysis \cite{downs1948variations} and Steiner analysis \cite{steiner1953cephalometrics}.
However, 
the accurate location of craniofacial landmarks from lateral cephalograms is a prerequisite for implementing these cephalometric analysis methods \cite{perillo2000effect}.
These landmarks are crucial since they serve as the basis for subsequent qualitative assessments of angles and distances.
Such assessments provide diagnostic information about a patient's craniofacial condition and influence treatment planning decisions \cite{legan1980soft,holdaway1983soft}.

\begin{figure*} [htb]
     \centering
     \begin{subfigure}[b]{0.30\textwidth}
         \centering
         \includegraphics[width=\textwidth]{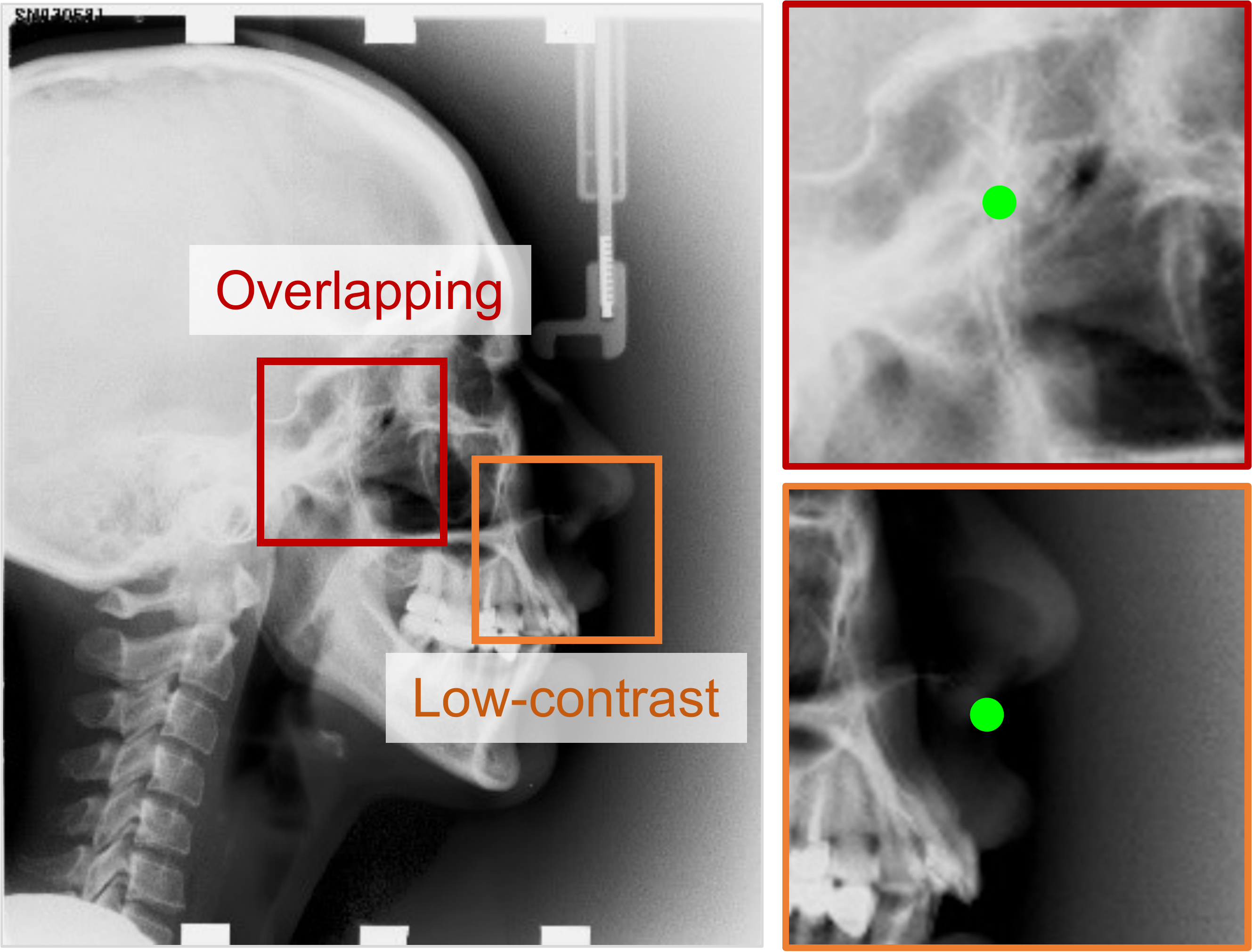}
         \caption{Structural overlap and low contrast}
         \label{fig:challenge_a}
     \end{subfigure}
     \hfill
     \begin{subfigure}[b]{0.37\textwidth}
         \centering
         \includegraphics[width=\textwidth]{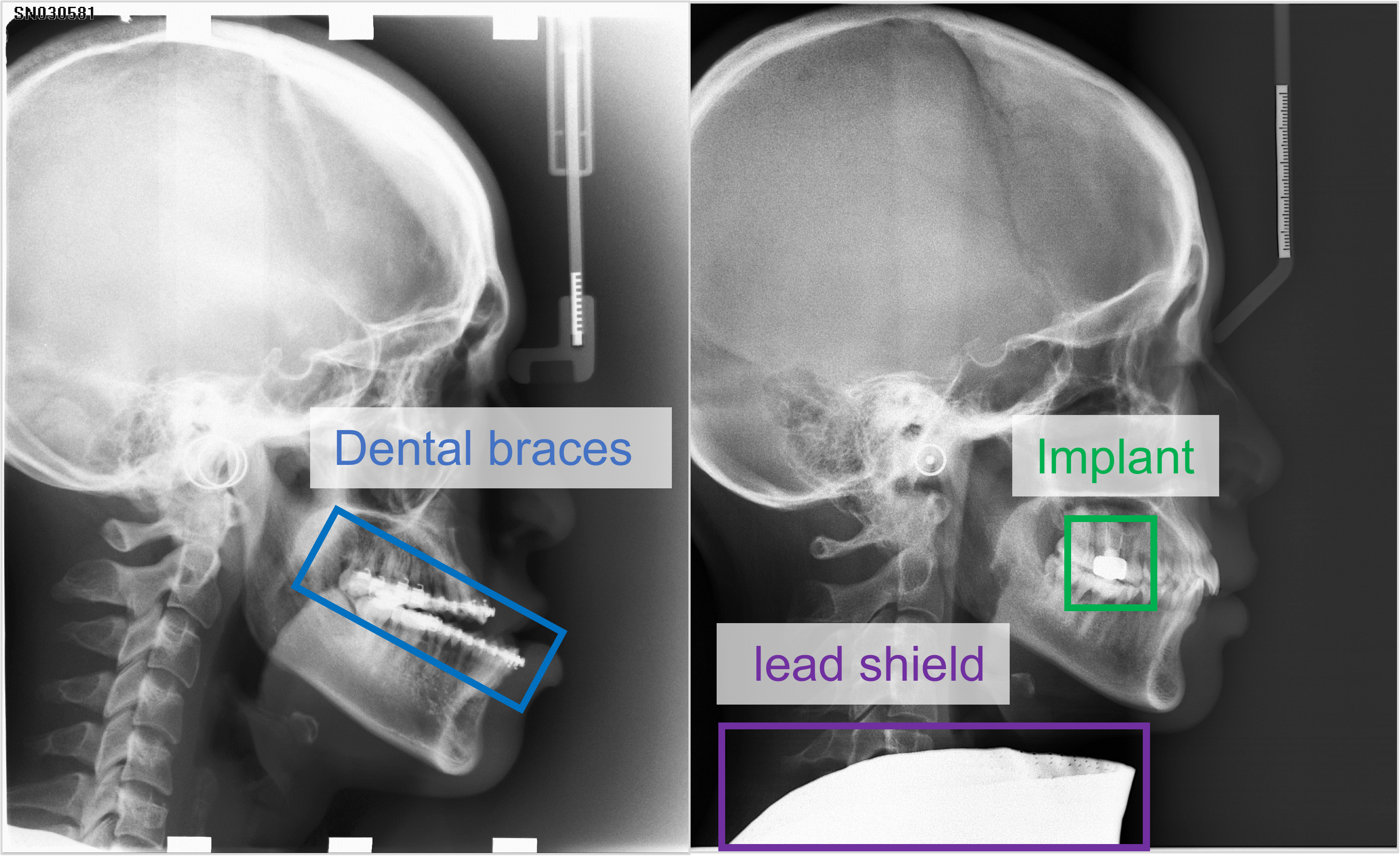}
         \caption{Landmark occlusion}
         \label{fig:challenge_b}
     \end{subfigure}
     \hfill
     \begin{subfigure}[b]{0.24\textwidth}
         \centering
         \includegraphics[width=\textwidth]{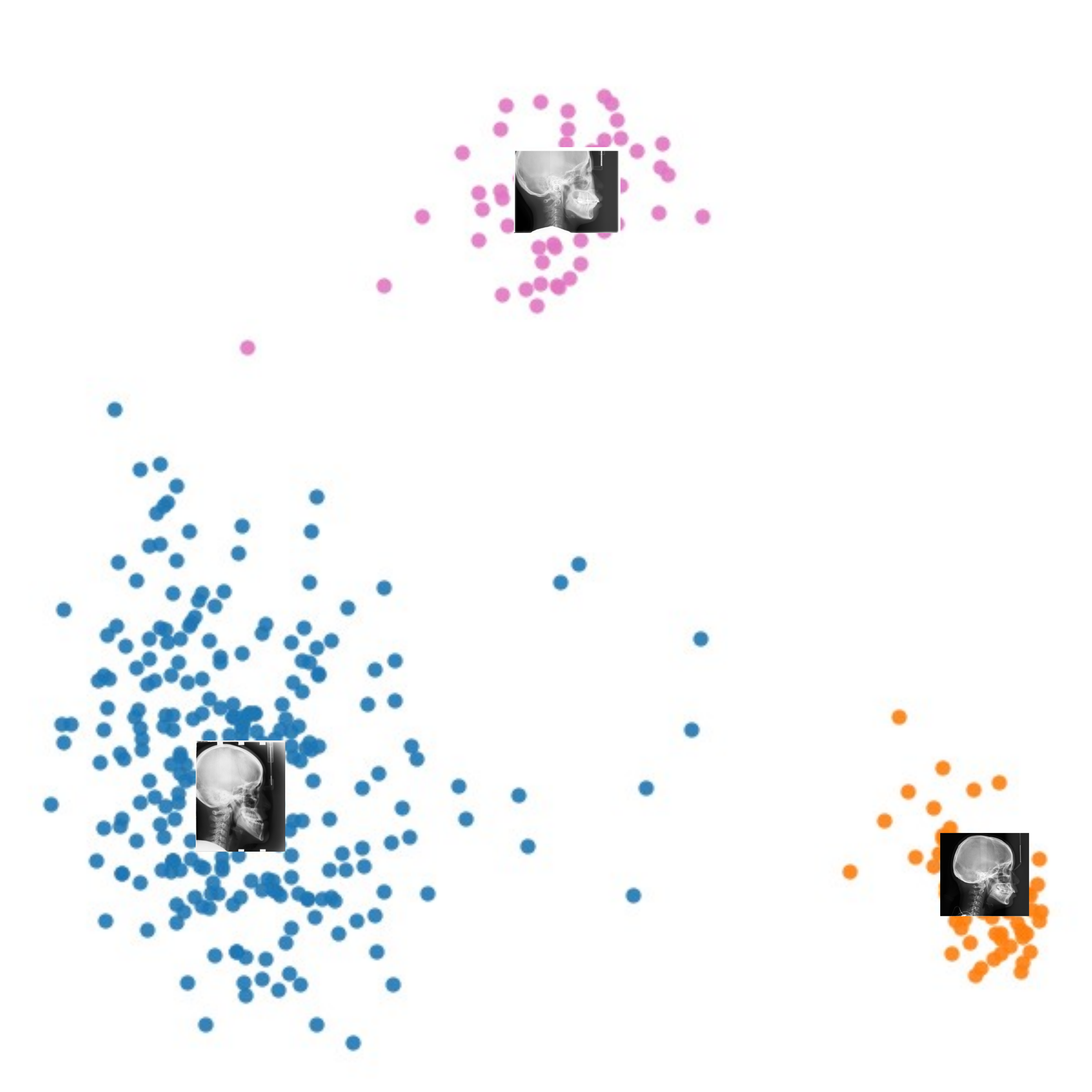}
         \caption{Site and individual variability}
         \label{fig:challenge_c}
     \end{subfigure}
    \caption{\textbf{Challenges in cephalometric landmark detection:}
    (a) Overlapping craniofacial structures (\textcolor[RGB]{255, 0, 0}{red} box) and  poor contrast (\textcolor[RGB]{233, 92, 22}{orange} box): In cephalometric X-ray images, craniofacial structures overlap, and soft tissue-related region is low-contrast, making it hard to distinguish individual components.
    (b) Landmark occlusion: Landmarks could be obscured by dental braces (\textcolor[RGB]{0, 0, 255}{blue} box), implants (\textcolor[RGB]{0, 255, 0}{green} box) or lead shields (\textcolor[RGB]{128, 0, 128}{purple} box).
    (c) Site and individual variability:
     t-SNE visualization of the challenge data from three medical centers reveals anatomical variations leading to differences in landmark appearance and location, not only across centers but also within the same center.
    }
    \label{fig:challenges}
\end{figure*}

However, accurate cephalometric landmark detection from lateral cephalograms remains challenging as illustrated in Fig. \ref{fig:challenges}:
\textbf{(1) Overlapping craniofacial structures:}
The inherent nature of X-ray imaging results in contralateral head structures appearing superimposed in a single image.
This complexity, as shown in Fig. \ref{fig:challenge_a},
makes it difficult to precisely identify and locate individual landmarks, particularly when craniofacial structures overlap \cite{mcclure2005reliability,delamare2010influence,sayinsu2007evaluation}.
\textbf{(2) Low-contrast:}
Soft tissue-related landmarks often suffer from low contrast,
rendering them indistinguishable from surrounding structures, and often leading to unreliable measurements \cite{mostafa2009soft,giannopoulou2020orthodontic}.
\textbf{(3) Landmark occlusion:}
As shown in Fig. \ref{fig:challenge_b},
there is occlusion of cephalometric landmarks, especially in the area of the teeth affected by dental braces and implants and in the neck area obscured by lead shields.
\textbf{(4) Site and individual variability:}
Different individuals naturally exhibit diverse appearances and geometric shapes of skull,
coupled with sex, age, and disease factors, which further lead to significant variances \cite{juneja2021review}.
Moreover, different scanners and imaging protocol settings across medical centers can introduce artifacts and image noise, further increasing imaging variability, as shown in Fig. \ref{fig:challenge_c}.

Manual marking of target landmarks is labor-intensive and time-consuming,
and suffers from intra- and inter-observer variability \cite{durao2015cephalometric}.
Therefore,
there has been a longstanding need for automatic and accurate landmark localization in clinical studies.
In recent years,
with the help of convolutional neural networks (CNNs) \cite{o2015introduction,lecun2015deep},
many cephalometric landmark detection methods have been proposed and
have achieved substantial progress in the era of deep learning. 
However,
as discussed in Section \ref{sec:relatedworks},
previous datasets only contain single-center, single-vendor data with a limited number of landmarks,
raising concerns about whether the performance achieved on these datasets can generalize to more diverse datasets.
Therefore,
it is worth considering that \textit{is cephalometric landmark detection truly a solved problem?}

To answer this question,
we first built a multi-center, multi-vendor and more comprehensive cephalometric landmark detection dataset,
named CL-Detection.
Then, based on this dataset,
we organized the \textbf{CL-Detection2023} challenge in conjunction with
the International Conference on Medical Image Computing and Computer Assisted Intervention (MICCAI) 2023.
The main topic of this challenge is to find automatic algorithms for accurately localizing cephalometric landmarks in lateral X-ray images.
Participants are required to develop automatic landmark detection algorithms,
and submit the algorithm Docker containers for evaluation on the test set to obtain the final ranking.
The CL-Detection2023 challenge provides a unique opportunity for participants from different backgrounds to compare their algorithms in an impartial manner.

In this paper,
we introduce a complete overview of the CL-Detection2023 challenge and discuss the top algorithms.
The main contributions are summarized as follows:
\begin{itemize}
\item We have constructed a comprehensive cephalometric landmark detection dataset, 
which provides a new benchmark for researchers to evaluate their new algorithms.

\item To answer the question \textit{'Is cephalometric landmark detection truely a solved problem?'},
we have analyzed the top-performing algorithms submitted to the challenge and summarized the results of the top teams.

\item We have presented various algorithmic techniques for improving
the accuracy of cephalometric landmark detection and provided insightful recommendations.

\item We have also investigated the current constraints of the existing solutions based on the
challenge submissions and identified areas where they fall short.
\end{itemize}

The rest of the paper is organized as follows.
In Section \ref{sec:relatedworks},
previous cephalometric datasets and cephalometric landmark detection methods are described.
In Section \ref{sec:description},
we detail the challenge organization, image datasets, annotation protocol, evaluation metrics and ranking scheme used within the challenge.
Then, we present a representative selection of methods which were submitted to the our challenge in Section \ref{sec:solutions}.
Next, we analyze the results obtained during the CL-Detection2023 challenge in Section \ref{sec:results} and finally draw analysis and conclusions in Section \ref{sec:discussion} and Section \ref{sec:conclusion}.

\section{Previous works}  \label{sec:relatedworks}
\subsection{Previous X-ray cephalometric datasets}
Two large datasets of clinical X-ray cephalometric data have been broadly accepted by the machine learning community in the last decade \cite{wang2015evaluation,wang2016benchmark,zeng2021cascaded}.
One dataset is our previous work, which was released in conjunction with the IEEE International Symposium on Biomedical Imaging (ISBI) challenge \cite{wang2015evaluation,wang2016benchmark}.
The ISBI challenge provides a database of 400 lateral X-ray images with 19 landmarks (250 for training, 50 for validation, and 100 for testing).
The outcome of the ISBI challenge revealed that the best scores were obtained by applying game theory and random forest method \cite{ibragimov2014automatic}.
It is noticed that, at that time, deep learning techniques were not yet popular, and their validation had not been explored.
In addition,
a more recent contribution to cephalometric research is the PKU cephalogram dataset introduced by \textit{Zeng et al.} in 2021 \cite{zeng2021cascaded}.
The dataset contains cephalograms of 102 patients, with each lateral X-ray image annotated with 19 cephalometric landmarks following the ISBI challenge rules.
The creation of PKU cephalogram dataset aimed to validate algorithm generalization,
highlighting the need for a diverse benchmark within the current research community.
Therefore, in the CL-Detection2023 challenge,
we extend our prior efforts \cite{wang2015evaluation,wang2016benchmark}
with multi-center, multi-vendor data and more landmark annotations,
which makes it possible to provide diverse data for benchmarking state-of-the-art (SOTA) methods.

\subsection{Non-deep learning methods}
The non-deep learning methods for cephalometric landmark detection can be divided into three main categories: (1) image filtering combined with knowledge-based landmark search \cite{levy1986knowledge,ren1998knowledge};
(2) model-based approaches \cite{hutton2000evaluation};
and (3) soft-computing approaches \cite{el2004automatic}.
However, These methods often struggle with accuracy due to anatomical variations and the quality of X-ray images \cite{leonardi2008automatic,kaur2015automatic}.
Furthermore,
in our previous work with the ISBI challenge \cite{wang2015evaluation,wang2016benchmark}, 
we used a unique dataset to evaluate and benchmark these non-deep learning techniques. However, a lack of standardized benchmarks for comparing deep learning methods remains a significant challenge in the field.

\subsection{Deep learning-based methods}
Existing deep learning-based methods can be classified into three categories: heatmap-based methods, coordinate-based methods and graph-based methods.
Heatmap-based methods \cite{oh2020deep,mccouat2022contour} model landmark locations as heatmaps and train deep neural networks to regress these heatmaps.
Coordinate-based methods \cite{gilmour2020locating,zeng2021cascaded} directly locate landmark coordinates from input images.
However,
both methods usually suffer from a major drawback of lacking a representation for shape,
which is important for the accurate detection of landmarks.
Graph-based methods \cite{li2020structured,lu2022landmark} can naturally model the structure of landmarks as a graph,
considering both landmark locations and the relationships between landmarks.
Besides, some recent works \cite{viriyasaranon2023anatomical} on medical landmark detection focus on transformer-based architectures.
These solutions can achieve a successful detection rate of over 75\% using a 2 mm precision range, which is close to expert level.
The results seem to answer the question we pose.
However, most of exiting method \cite{payer2016regressing,zeng2021cascaded,li2020structured} have been only evaluated on a single-source dataset,
and further verification of the generalization of these methods on a diverse dataset is needed.

\begin{figure*}[!t]
\centering
\includegraphics[width=1\textwidth]{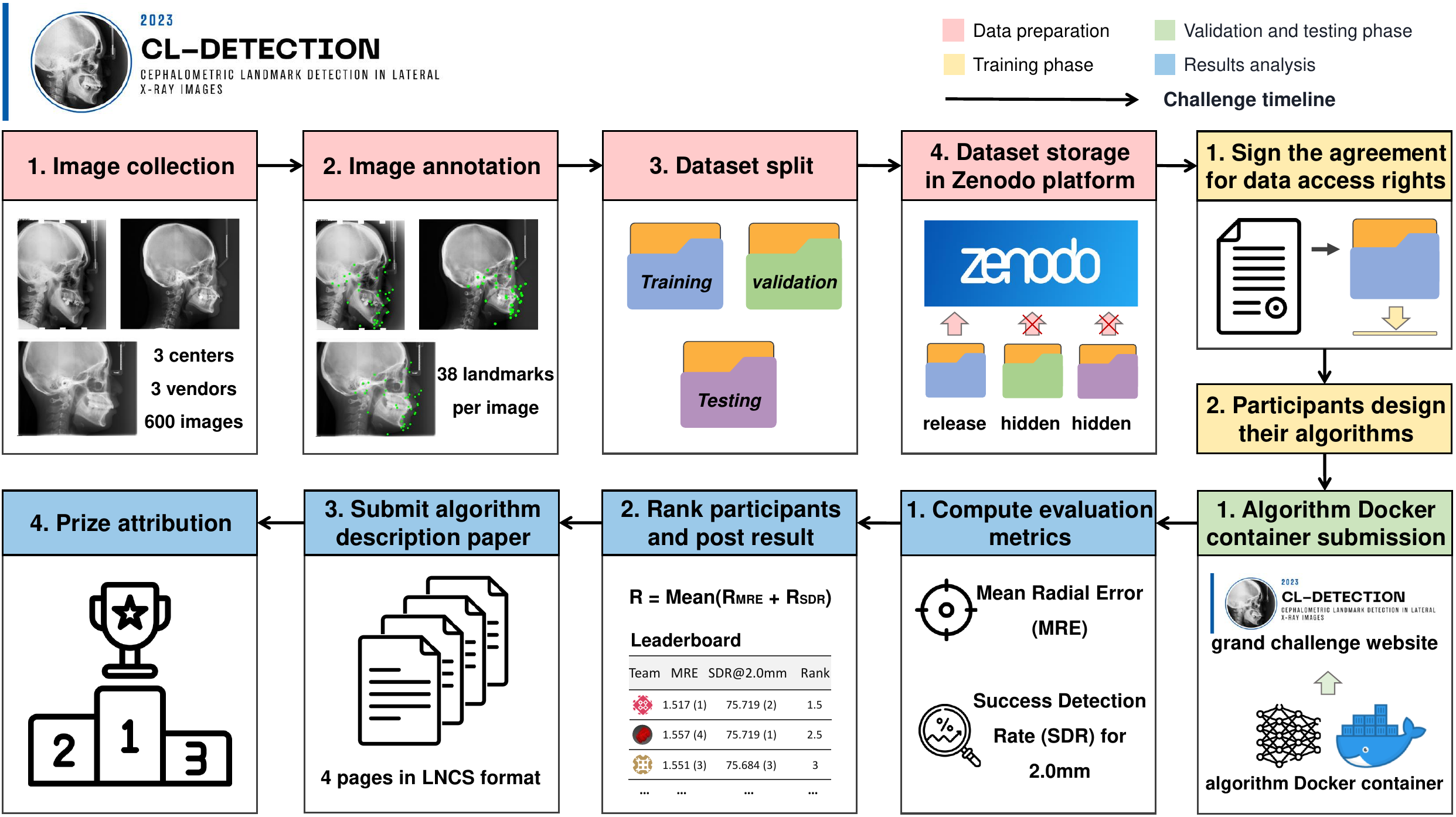}
\caption{\textbf{The workflow of the MICCAI CL-Detection2023 challenge consists of four stages:} (1) Data preparation, (2) Training phase, (3) Validation and testing phase, and (4) Result analysis.}
\label{fig:challenge_workflow}
\end{figure*}

\section{Challenge description} \label{sec:description}

\subsection{Organization}
When organizing the CL-Detection2023 challenge and writing this paper,
we followed the BIAS guideline \cite{maier2020bias}.
The challenge was structured into four stages, providing a well-structured workflow to enhance the success of the initiative, as shown in Fig. \ref{fig:challenge_workflow}.
Specifically,
the challenge was a collaborative effort of different institutions and researchers from Shenzhen University General Hospital, Shenzhen University, and National Taiwan University of Science and Technology.
Meanwhile,
our challenge was part of the Dental Enumeration and Landmark Detection Techniques Advancement (DELTA) workshop.
The CL-Detection2023 challenge was located on the Grand-Challenge platform\footnote{\url{https://grand-challenge.org/}} and the training dataset was hosted on the Zenodo platform\footnote{\url{https://zenodo.org/}}.
Participants were required to first sign the challenge rule agreement and send it to the official mailbox to get access to the training set data.
Everyone was encouraged to enter the competition, but 
members of the challenge organizer's institutes were not eligible for awards.

Our CL-Detection2023 challenge included three phases, a training phase, a validation phase and a testing phase.
During the training phase,
participants could develop a full-automatic detection algorithm with training images and corresponding annotations after their registration applications were approved.
In the validation phase,
to prevent over-fitting, 
participants were allowed to upload a maximum of five algorithm Docker containers to the official website for validation.
Scores and rankings were automatically calculated and promptly shown upon submission of Docker containers.
During the testing phase,
participants could upload up to two algorithm Docker containers,
but only one working Docker container per team was confirmed to produce the final test results.

To assist participants in model construction and algorithm Docker submissions,
we provided several baseline models. In particular,
we offered RetinaNet \cite{lin2017focal} based on the MMDetection object detection framework \cite{chen2019mmdetection},
and HRNet \cite{wang2020deep} based on the MMPose landmark location framework \cite{contributors2020openmmlab}, as two primary baselines.
In addition,
we provided a U-Net \cite{ronneberger2015u} baseline model for heatmap prediction based on the pure PyTorch framework \cite{paszke2019pytorch},
which can allow participants to seamlessly integrate and utilize it.
All three baseline models came with detailed documentation and a comprehensive algorithm container submission process for immediate use.

Besides, for a fair comparison,
the data used to train algorithms were restricted to those provided by this challenge.
Pre-trained models from ISBI challenges or PKU cephalogram dataset were also not allowed to be used in the challenge.
The top three performing teams were awarded with certificates and 500 euros each.
Their final results were announced publicly on challenge website.
Besides,
the top 10 teams were invited to show their excellent algorithms at the MICCAI workshop and to be co-authors of the challenge review paper.
After challenge, participating teams could publish their own result separately, however, it needs to obey the citation rules.

The CL-Detection2023 challenge was one-time event with fixed conference deadlines.
However,
the challenge submission system and dataset will remain available also after the first evaluation in the context of MICCAI 2023.
Our challenge website was first made publicly available on the March 15, 2023.
Participants could register until April 1, 2023,
and had four months to submit their entries until the deadline for submissions on August 16, 2023.
The training data was released on May 1, 2023,
and results were announced on August 30, 2023, half a month after the submission deadline.

\begin{table*}[!t]
  \centering
  \caption{\textbf{The three image dataset information included in the CL-Detection challenge},
  as well as the corresponding division of training, validation and testing sets.}
  \label{tab:data}
  \setlength{\tabcolsep}{3.5mm}
  \renewcommand\arraystretch{1.2} 
  \begin{tabular}{ccccccc}
    \hline \toprule
    & \multirow{2}{*}{Number of samples} & \multicolumn{3}{c}{Dataset division} & \multirow{2}{*}{Image size}  & \multirow{2}{*}{Pixel spacing (mm)}
    \\ \cmidrule(r){3-5} 
    &   & Training & Validation & Testing   &   &  \\ \hline
    ISBI challenge                     & 400   & 266   & 30    & 104 & 1935 $\times$ 2400  &  0.100  \\
    PKU cephalogram dataset            & 102   & 68   & 12    & 22  & 2089 $\times$ 1937  &  0.096  \\
    Shenzhen General Hospital dataset  & 98    & 66   & 8      & 24  & 2880 $\times$ 2304  &  0.125  \\ 
    \hline \toprule
  \end{tabular}
\end{table*}

\begin{figure*}[!t]
\centering
\includegraphics[width=1\textwidth]{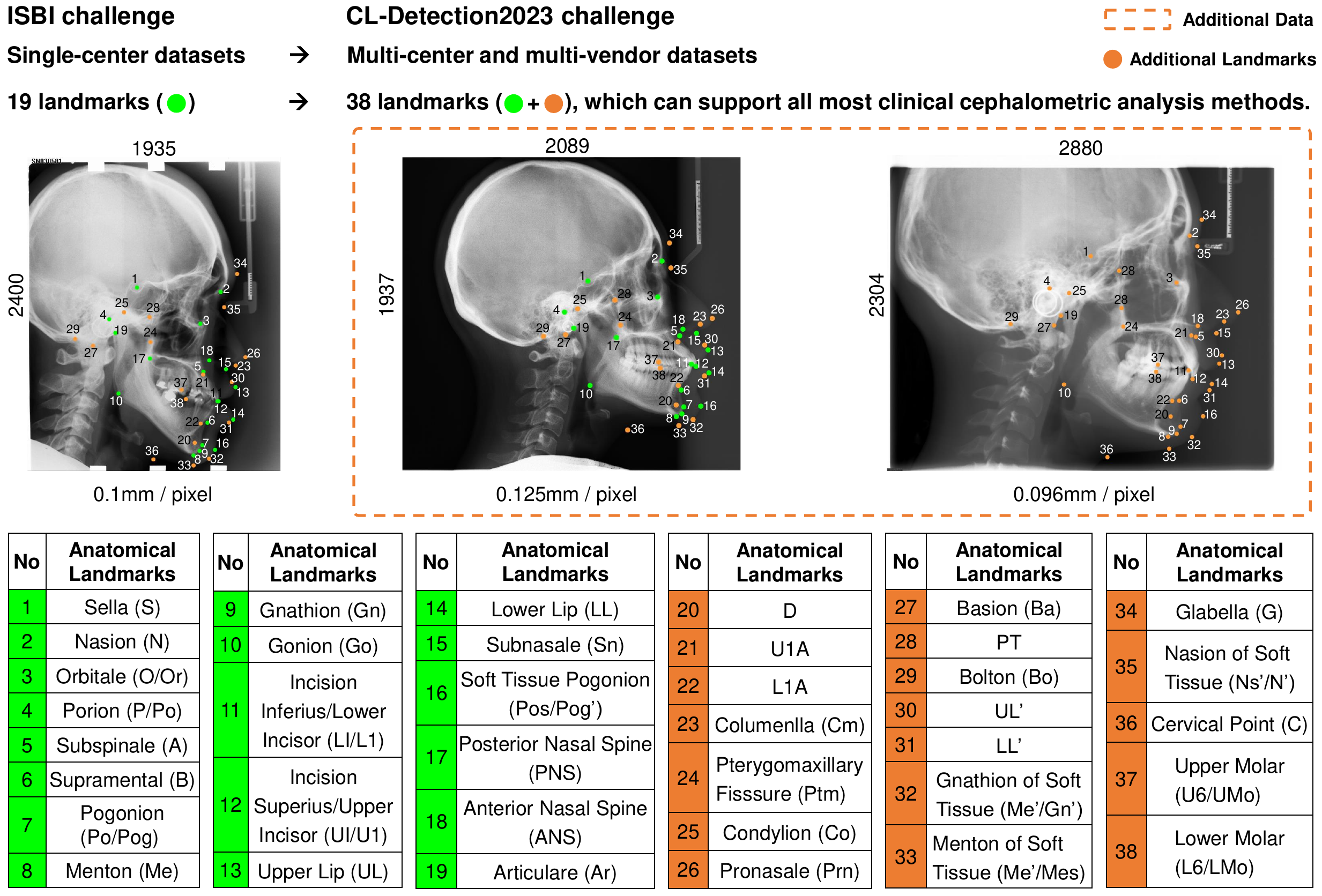}
\caption{\textbf{Comparative analysis of the differences between our previous work ISBI challenge and 
CL-Detection2023 challenge.}
The green highlights represent the anatomical landmarks featured in the ISBI 2015 challenge.
In CL-Detection2023 challenge, we have extended dataset from a single center to a multi-center, multi-vendor and more landmark annotations, which are highlighted in orange.
}
\label{fig:challenge_landmarks}
\end{figure*}

\subsection{Image datasets}
The CL-Detection2023 challenge aims to provide a dataset that better reflects real-world applications 
and includes a more diverse range of lateral X-ray cases.
The challenge cohort consists of subjects with cephalometric analysis,
with the hope that the developed algorithm could be adaptable to a broader patient population from any hospital or medical center.
The dataset source, division and detailed parameters are shown in Table \ref{tab:data}.
In particular, to include more diverse cases, our dataset consists of 600 2D lateral X-ray images from two existing datasets:
ISBI challenge (400 cases) \cite{wang2015evaluation,wang2016benchmark}, PKU cephalogram dataset (102 cases) \cite{zeng2021cascaded} and a new dataset from Shenzhen General Hospital (92 cases). 
The first public 98 dental X-ray images collection and use has been approved by the Research Ethics Committee of Shenzhen University General Hospital.
ISBI challenge and PKU cephalogram dataset are licensed under the Creative Commons license CC-BY-SA 4.0.
Under the license,
we are allowed to modify the datasets and share or redistribute them in any format.
For the CL-Detection2023 challenge, 
all the data have been anonymized and is compliant with the CC BY-NC-ND (Attribution-NonCommercial-NoDerivs) license. 

As shown in Fig. \ref{fig:challenge_landmarks} and Table \ref{tab:data}, 
compared with our previous ISBI challenge,
the challenge dataset includes data from three centers
and three different acquisition devices.
Then, all the images were shuffled and divided into three sets, i.e., the training, validation and testing sets with case number 400, 50 and 150, respectively.
Training and test cases both represent cephalometric X-ray image of each patient.
The training cases include the corresponding annotations of landmark.
A case refers to a cephalometric patient.
We did not separate the dataset according to the proportion of data from each center to avoid over-fitting to any particular center.

\subsection{Reference detection and annotation protocol}
The expert references are manually-locate anatomical landmarks for 38 bone-related and soft-tissue structures in lateral X-ray images.
Fig. \ref{fig:challenge_landmarks} provides visual representation of the locations of these craniofacial landmarks.
The annotation work involved three experts with more than five-year clinical experience.
Initially, the annotation process was carried out by two senior doctors.
The doctors extended the existing benchmark datasets with more landmark annotations,
and annotated landmarks from scratch on the first public dataset of 98 images.
Next, they conducted a double-check on all landmarks and had to reach consensus in case of discordance.
Finally, one senior doctor with more than 20-years experience verified and refined the annotations.

The annotation rules followed the the Contemporary Orthodontics guidelines \cite{ricketts1982orthodontic}.
All 38 cephalometric landmarks must be annotated based on their anatomical definitions,
and the corresponding landmarks need to satisfy the defined positional relationships.
For example, the Sella (S) landmark should correspond to the midpoint of the sella turcica.
Majority of the landmarks are clear enough to allow accurate annotation.
The main difficulty when annotating landmarks is correctly identifying low-contrast soft tissue edges.
These edges serve as crucial image references for locating soft tissue-related landmarks.
As such,
doctors were allowed to adjust the contrast and brightness of the images if necessary to improve visual interpretation during annotation.
In addition,
we have calculated the inter-observer variability of the two senior doctors with over five years of experience on a subset of 50 cases to assess human performance.
Current findings suggest that Inter-class Correlation Coefficients (ICCs) \cite{bartko1966intraclass,shrout1979intraclass} are excellent for all landmarks (ICCs over 0.90 for all landmarks).
Details on inter-observer variability for each landmark are in Appendix A.
In addition, our double-check approach further ensures the reliability of the annotations in our dataset.
Finally,
the ground truth landmarks are stored in JSON format \cite{carter2018understanding} files.
Participants could access the training images with corresponding annotations.
The images and annotations for the validation and testing set are held by the organizers.

\subsection{Evaluation metrics and and ranking scheme}
The methods developed by the participants were evaluated on test sets for which the ground-truth was hidden.
In order to evaluate the algorithm methods in a fair and reproducible manner,
we adopted the Mean Radial Error (MRE) and the Success Detection Rate (SDR) for 2.0 mm (SDR@2.0mm) as evaluation metrics. 
These two metircs are often used in landmark detection applications \cite{wang2016benchmark,zhong2019attention,li2020structured,chen2019cephalometric}.
The MRE metric measures the difference between two landmarks.
The MRE is formulated as follows:

\begin{equation}
    MRE = \frac{\sum_{i=1}^N \sqrt{\Delta x_i^2 + \Delta y_i^2}}{N}
\end{equation}
where $\Delta x_i$ and $\Delta y_i$ are the absolute distances between the obtained and reference landmarks in the $x$ and $y$ directions, respectively,
and $N$ is the number of detection landmark. 

Furthermore,
SDR@2.0mm metric measures the accuracy between the ground truth and the automatic results,
which provides a good measure of localization quality.
The SDR with precision less than 2.0mm is formulated as follows:
\begin{equation}
    p_z = \frac{\#\{j: ||L_d(j) - L_r(j)|| < z\}}{\#\Omega} \times 100\%
\end{equation}
where $L_d$ and $L_r$ represent the locations of the detection landmark and the referenced landmark, respectively;
z denotes 2.0mm precision range used in the evaluation;
and $\#\Omega$ represents the number of detection landmarks.

For each test case, we calculated the MRE and the SDR@2.0mm between the ground truth and the participants' results. 
Besides,
we have excluded the participants who fail to report on the whole testing set.
After we got all the teams’ predictions, we implemented the following ranking scheme:

\begin{itemize}
\item Step 1. We took the mean of the MRE and the mean of the SDR@2.0mm over the test cases. 

\item Step 2. The mean MRE and mean SDR@2.0mm were ranked separately among the teams.
MRE is ranked in descending order and SDR@2.0mm is ranked in ascending order.

\item Step 3. After we got all the team's metrics ranks,
a final rank was given by taking the average of the two ranks.

\item Step 4. Based on the rankings in step 3, the final
ranking was determined.
In the case of equal average rankings for two teams, they were considered tied.

\end{itemize}

All the submitted algorithm Docker containers were evaluated on a Amazon's ml.m5.large instance, equipped with a maximum GPU memory of 32GB, 2 CPU cores, and 8GB of RAM.
The allowed processing time for 50 test images was limited to 10 minutes,
including model loading, I/O, preprocessing, and inference.
If the GPU overflows or time limit was exceeded,
the submission were considered a failure.
To make a transparent challenge, the evaluation code was released on the challenge evaluation page.

\begin{figure} [t]
\centering
\includegraphics[width=\linewidth]{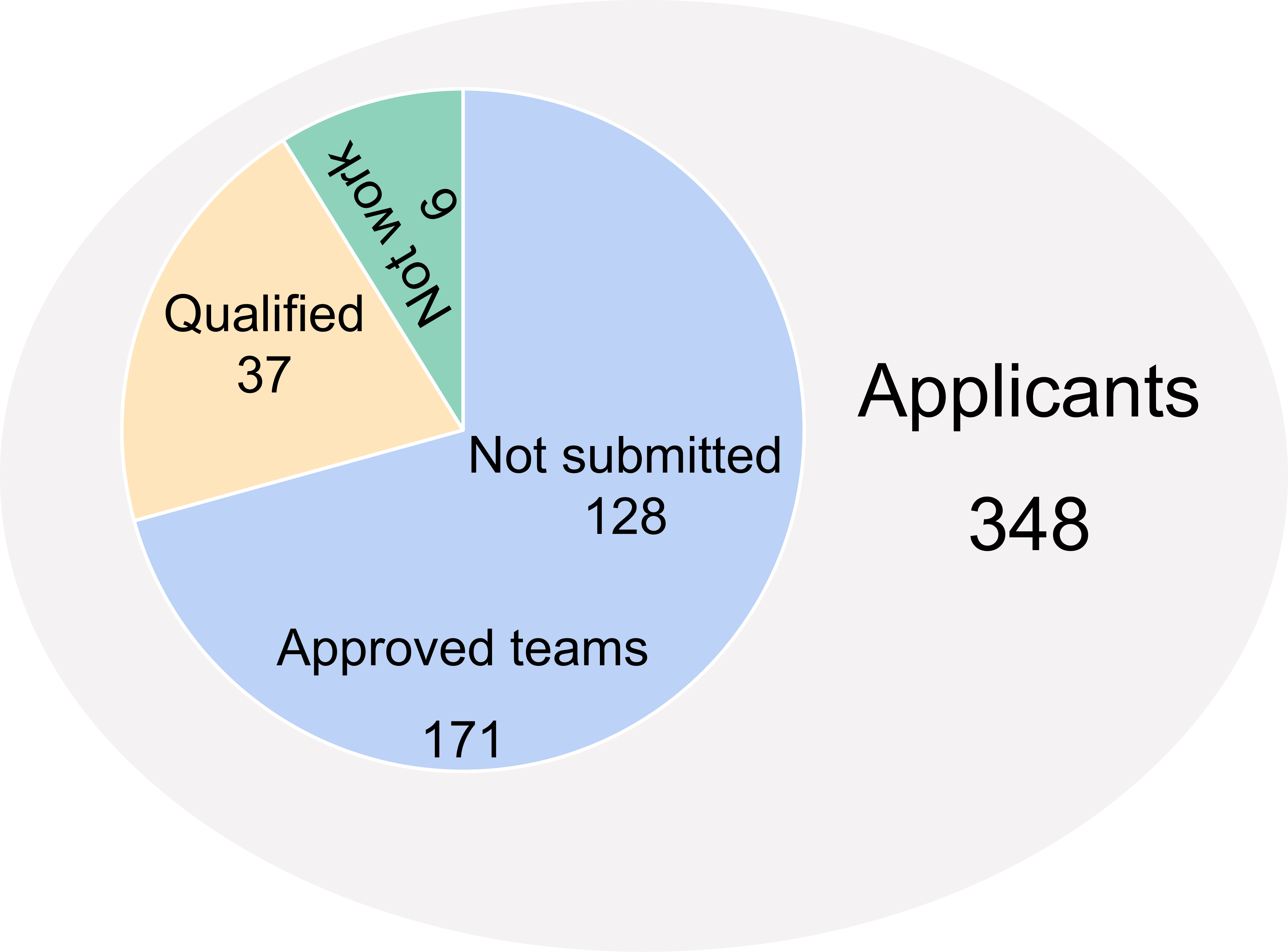}
\caption{\textbf{Summary of CL-Detection challenge participants and submissions.}
There were 348 teams registering on the official grand-challenge website and
171 of them were approved before the end of the training phase. Finally, 46
teams submitted validation results and 37 teams submitted Docker containers
for test leaderboard.
}
\label{fig:submission_information}
\end{figure}

\section{Competing solutions} \label{sec:solutions}

This section provides a comprehensive description of the participating approaches.
Fig. \ref{fig:submission_information} shows information about participants and submissions. 
Specifically,
we received more than 300 applications from over 30 countries on the grand-challenge webpage
and 171 teams were approved.
During the validation phase,
46 teams submitted validation results, but 4 Docker containers can not work.
Finally, during the testing phase, 37 teams submitted Docker containers with 77 qualified results,
and an additional 6 teams had 7 Docker container submissions that failed to execute.
We highlight the main features of the top ten teams in Table \ref{tab:top10}.
Further details of their algorithms are outlined as below.

\begin{table*} [!t]
  \centering
  \caption{\textbf{Summary of the benchmark methods of top ten teams.}
  It provides a brief description of the different frameworks, networks, and key strategies utilized by each team to enhance their performance.}
  \label{tab:top10}
  \setlength{\tabcolsep}{7mm} 
  \renewcommand\arraystretch{1.25} 
  {\begin{tabular}[]{llll}
    \hline
    \toprule
    Team                 & Framework                  & Network                                     & Highlighting methods                                                          \\ \hline
    \multirow{4}{*}{CTT} & \multirow{4}{*}{Two-stage} & \multirow{4}{*}{U-Net with Efficient-B3 encoder} & Sigma head to learn the variance for landmarks; \\ 
            & & & Large heatmap output\\
            & & & L1 loss;\\ 
            & & & Models ensemble.\\ \hline
    \multirow{4}{*}{SUTD}  & \multirow{4}{*}{One-stage} & \multirow{4}{*}{HRNet}                    & Generated large heatmap prediction; \\
            & & &  Super-resolution head; \\
            & & &  Deep supervision. \\ 
            & & &  Test-time augmentation. \\ \hline
    \multirow{3}{*}{WHU} & \multirow{3}{*}{One-stage} & \multirow{3}{*}{HRNet} & Various data augmentation techniques; \\
            & & & Large heatmap output; \\
            & & & Models ensemble. \\ \hline
    \multirow{3}{*}{PKU} & \multirow{3}{*}{One-stage} & \multirow{3}{*}{Up-HRNet}       & Reverse pyramid feature fusion strategy;\\
            & & & Histogram equalization; \\ 
            & & & Point-shift strategy to overcome the annotation noise. \\ \hline
    \multirow{3}{*}{XDU}  & \multirow{3}{*}{One-stage} & \multirow{3}{*}{U-Net}                      &  Local and global information integration; \\  
            & & & Dilation convolutions; \\ 
            & & & Models ensemble.  \\ \hline
    \multirow{3}{*}{UM} & \multirow{3}{*}{One-stage} & \multirow{3}{*}{HRNet}                 & Various data augmentation techniques; \\  
            & & & Multi-scale result fusion; \\
            & & & Models ensemble.      \\ \hline
    \multirow{2}{*}{OUC} & \multirow{2}{*}{Two-stage} & \multirow{2}{*}{HRNet}                    & Suitable data augmentations; \\  
            & & & Parallel refine network; \\ \hline
    \multirow{2}{*}{DGMIF} & \multirow{2}{*}{One-stage} & \multirow{2}{*}{U-Net}                      & Affine and intensity change data augmentation;  \\  
            & & & Models ensemble. \\ \hline
    SCNU              & One-stage                  & VGG19                                       &  Attention feature pyramid fusion module.                                                      \\ \hline
    CHZUSM            & One-stage                  & HRNet, U-Net, VNet                          & Models ensemble.                                                   \\ \hline \toprule
  \end{tabular}}
\end{table*}

\subsubsection{T1: Chohotech Technology Co., Ltd (CTT)}
This submission was made by Chenglong Ma and Feihong Shen. 
The authors developed a two-stage cephalometric landmark detection framework that operates in a coarse-to-fine manner.
In the training phase, they designed two aspects (i.e model and loss function) to achieve SOTA performance.
The primary idea of their method is to maintain the highest possible image resolution.
For the model, 
they modified the U-Net model \cite{ronneberger2015u} with Efficient-B3 encoder to perform landmark detection.
The backbone network consists of a 5-depth encoder, a 4-depth decoder, and a Sigma head with multi-layer perceptron.
Specifically,
in the coarse stage, 
the original image was resized to $1024 \times 1024$ and input into the network, generating a heatmap prediction of size $512 \times 512$.
After obtaining the coarse result,
the region corresponding to the location of the landmarks was cropped out.
These cropped region was then resized to a size of $2048 \times 2048$ and input into the network structure of the first stage.
The output heatmap from this stage has a size of $1024 \times 1024$.
To address the issue of high data imbalance in the second stage,
the framework adopted a simple strategy of cropping a $256 \times 256$ patch centered around the coarse landmark positions to calculate the loss.
For the loss, 
CTT applied the L1 loss to the predicted heatmap and landmark locations through a soft-argmax process \cite{luvizon2019human}.
The ground truth heatmap was generated based on the standard deviation obtained from the Sigma head regression. 
In addition,
CTT employed a 10-fold cross-validation approach to evaluate the performance of their models. Afterward, they identified the four top-performing models based on their performance on the validation sets.
For the final test set results,
a model ensemble technique was utilized, involving the averaging of the outputs from the four selected models.

\subsubsection{T2: Singapore University of Technology and Design (SUTD)}
This submission was made by Qian Wu, Si Yong Yeo, and Jun Liu.
The authors modified HRNet \cite{wang2020deep} with multiple aspects.
(1) Feature fusion:
improve the feature fusion of the original HRNet with separable convolutions \cite{chollet2017xception,howard2017mobilenets} to efficiently aggregate features from different levels.
The fuse module operated in the following steps:
first, the features were concatenated after passing through a convolution block and an upsampling operation.
Next, the concatenated features underwent a point-wise convolution (PW Conv) block.
Then, two consecutive modules were applied to further aggregate features.
Finally, a PW Conv operation produced the fused result.
(2) Generate high-resolution heatmap:
to mitigate quantization deviations by bridging the resolution gap between the input image and the predicted heatmaps,
SUTD adopted efficient convolutions and parameter-free pixel shuffle operation for upsampling high-resolution heatmaps.
(3) Deep supervision:
the deep supervision scheme was implemented for the supervision of two-scale heatmaps to improve the discriminability of the model’s extracted features.
In addition,
during the testing phase, 
left-right flipped test time augmentation (TTA) was used to further improve the model’s performance.
In the post-processing stage,
the DARK \cite{zhang2020distribution} method was incorporated as their debiasing approach.

\subsubsection{T3: Wuhan University (WHU)}
This submission was made by Xianan Cui and Xianzheng Ma.
WHU believed that higher resolution potentially leads to improved performance.
Therefore, they opted to utilize HRNet \cite{wang2020deep} as backbone network, which is designed for high-resolution learning.
Specifically,
WHU designed an approach to upsample the heatmap to the original resolution of the input image through deconvolution blocks.
Each deconvolution block consists of two consecutive deconvolution layers with 128 channels each to upscale the heatmaps to the original resolution.
During training phase,
they employed various data augmentation techniques, including random affine transformations, random Gaussian blur, random dropout, and horizontal flipping,
to better generalize the CL-Detection data.
In the inference phase, WHU utilized flipping as a self-ensemble strategy.
For post-processing, WHU adopt DARK \cite{zhang2020distribution} as their debiasing method.
To get better results for the challenge,
they ensembled four best models by average for the final submission.
Prior to being fed into the model,
all X-ray images were resized to a resolution of $1024 \times 1024$.

\subsubsection{T4:  Peking University (PKU)}
This submission was made by Kunlun Xu and Tao Zhang.
PKU proposed a novel architecture termed the Reverse Pyramid Network (RPNet) that used HRNet backbone \cite{wang2020deep} with an upper-level sampling layer.
The network architecture was characterized by multiple consecutive stages.
Within each stage,
the input image undergoes downsampling, HRNet network and upsampling to acquire the resulting output.
Upon the completion of each stage, the output comprises both a feature map and a landmark heatmap, both aligned with the designated output resolution.
To enhance the visibility of skin tissue features and mitigate this dynamic range,
the local histogram equalization with a mesh size of 32 was implemented to enhance image before data augmentation.
Beisdes,
PKU incorporated various common data augmentation techniques, including random cropping, random scaling, random rotation, random blurring, random noise injection, color enhancement, and flipping, To enhance the visibility of skin tissue features and mitigate this dynamic range, to improve the algorithm’s performance.
During the testing phase,
left-right flipped TTA was used to address the issue of landmark deviation randomness.

\subsubsection{T5: Xidian University (XDU)}
This submission was made by Pengfei He and Jiale Zhang.
XDU modified the U-Net model \cite{ronneberger2015u} with PVT\_v2 backbone \cite{wang2022pvt} to capture global spatial structure information.
In addition, a fusion module called the local-to-global consistency spatial structure awareness module was developed to enhance the integration of local and global information.
In this module,
four outputs from the U-Net's upsampling layer were concatenated to create a combined feature map. 
Then, a series of convolutions,
including traditional convolutions with 1 $\times$ 1 and 3 $\times$ 3 kernel sizes, as well as two dilation convolutions \cite{yu2015multi} with a 3 $\times$ 3 kernel and expansion rates of 2 and 4, 
were used to capture information at different scales.
The whole model was trained with Adaptive Wing loss \cite{wang2019adaptive} to quickly locate the landmarks.
Due to the significant error of a single model,
XDU used an average method to combine the outputs of eight models for the final ensemble.

\subsubsection{T6: University of Macau (UM)}
This submission was made by Dongqian Guo He and Wencheng Han.
UM leveraged the HRNet \cite{wang2020deep} to create multiple models and ensembled their predictions to predictions to produce final outputs.
Specifically, they trained five models with input image resolutions of 800 $\times$ 800, 1024 $\times$ 1024, 1280 $\times$ 1280, and 1408 $\times$ 1408 (including two models at 800 $\times$ 800),
incorporating random cropping and shifting for data augmentation.
Furthermore, two distinct models were trained, one with and one without random brightness augmentation, both operating at a resolution of 512 $\times$ 512. 
In summary, a total of seven models with varying resolutions and
data augmentation techniques were trained for prediction.
Among the seven prediction results
for each landmark, we discarded the one with the lowest confidence.
The remaining six prediction results were then averaged to obtain the final predicted
coordinates.

\subsubsection{T7: Ocean University of China (OUC)}
This submission was made by Chenglong Ma.
T7 selected HRNet \cite{wang2020deep} for extracting multi-scale information from cephalometric lateral radiography images.
It employed $1 \times 1$ convolutions to harmonize the channel dimensions with high-resolution features,
followed by upsampling operations to bring features to a same size for fusion at different levels.
This fusion of feature information was achieved by addition operation.
At the same time, a refinement sub-network used the low-resolution feature map extracted by the HRNet in the first stage to further adjust the landmark positions.
Both stages of the network used the mean square error loss function to calculate the heatmap loss predicted by the model to optimize the weights.

\subsubsection{T8: Daegu-Gyeongbuk Medical Innovation Foundation (DGMIF)}
This submission was made by Hyunseok Lee, Hyeonseong Hwang, Yeonju Jeong, Seungyong Han, Gyu-sung Ham and Kanghan Oh.
DGMIF proposed a BigU-Net that used U-Net backbone \cite{ronneberger2015u}, but distinguish from the standard U-Net.
They adopted instance normalization \cite{ulyanov2016instance} instead of batch normalization as the former was more friendly to small batch size.
In order to response to large image inputs, 
the number of output channels of the first and last layers of U-Net were increased from
64 to 128.
The heatmap was conceptualized as a Laplacian distribution, supervised by the L2 loss function. 
DGMIF trained different versions of U-Net, including different input image sizes, U-Net types, and  the number of power for heatmap.
Then, the precise landmark location was obtained through model ensemble by averaging predicted heatmaps from each model.

\subsubsection{T9: South China Normal University (SCNU)}
This submission was made by Xiaotong Xie, Zhangnan Zhong, and Yaheng Fan.
To extract semantically enhanced features and elevate model detection capabilities, 
SCNU used the VGG19 network \cite{simonyan2014very} as the encoder of U-Net, and incorporated the Attention Feature Pyramid Fusion module (AFPF) \cite{chen2019cephalometric} into both the encoder and decoder of U-Net.
The model produced two channels, heatmaps and coordinate offsets.
For heatmap supervision,
dice loss and cross-entropy loss were employed,
while the L1 loss function was applied for offset map supervision,
contributing to a further enhancement in detection accuracy.
In all experimental configurations, all image were resized to $640 \times 800$, and a radius of 31 was employed in the generation of the heatmap.

\subsubsection{T10: Children’s Hospital Zhejiang University School of Medicine (CHZUSM)}
This submission was made by Zhu Zhu, Liuling Dong, Yu Liu, Zhihao Yang and  Xiaoling Gu.
The authors adopted an ensemble-based approach that leverages multiple deep learning models to address the cephalometric landmark detection task.
In particular, three models (U-Net \cite{ronneberger2015u}, VNet \cite{milletari2016v}, and HRNet \cite{wang2020deep}) were employed.
Each image was resized to 1024 $\times$ 1024 to maintain standardized height and width for feeding into the network.
All three networks were collectively trained with 38 landmarks, producing 38 heatmaps as output.
Besides, a weighted average ensemble of these models was crafted to derive the final set of 38 heatmaps,
with a primary focus on minimizing the MRE value.
During the training phase,
an early stopping mechanism with patient epoch 20 was implemented to prevent over-fitting.

\begin{table*}[!t]
  \centering
  \caption{\textbf{Quantitative evaluation and ranking results of the top 10 teams in terms of (mean ± standard deviation) MRE, SDR@2.0mm.} The arrows indicate which direction is better.}
  \setlength{\tabcolsep}{5.2mm} 
  \renewcommand\arraystretch{1.2} 
  
  \label{tab:all_results}
  \begin{tabular}{lccccc}
  \hline
  \toprule
    \multirow{2}{*}{Teams} &
    Mean Radial Error &
    \multirow{2}{*}{MRE Rank} &
    Success Detection Rate  &
    \multirow{2}{*}{SDR@2.0mm Rank} &
    Mean Rank \\
    & (MRE) (mm) $\downarrow$ &   & (SDR) 2.0mm (\%) $\uparrow$ &  & (Overall Rank) \\
    \hline
    T1 & $1.518\pm1.620$  &  1  & $75.719\pm\phantom{0}9.847$ &  1  &  1~~ (1)   \\
    T2 & $1.556\pm1.844$  &  3  & $75.719\pm10.097$           &  1  &  2~~ (2)   \\
    T3 & $1.551\pm1.868$  &  2  & $75.684\pm10.159$           &  3  &  2.5 (3)   \\
    T4 & $1.584\pm1.748$  &  4  & $75.211\pm10.346$           &  5  &  4.5 (4)   \\
    T5 & $1.621\pm2.176$  &  7  & $75.526\pm10.236$           &  4  &  5.5 (5)   \\
    T6 & $1.620\pm1.834$  &  6  & $74.175\pm\phantom{0}9.697$ &  6  &  6~~ (6)   \\
    T7 & $1.616\pm1.701$  &  5  & $73.596\pm\phantom{0}9.942$ &  8  &  6.5 (7)   \\
    T8 & $1.631\pm1.724$  &  8  & $73.772\pm10.043$           &  7  &  7.5 (8)   \\
    T9 & $1.661\pm1.792$  &  9  & $73.421\pm10.450$           &  9  &  9~~ (9)   \\
    T10 & $1.676\pm1.744$ & 10  & $73.456\pm\phantom{0}9.512$ & 10  &  10~ (10)  \\ \hline
    Baseline \cite{wang2020deep} & $2.588\pm6.805$ &  -  &  $65.404\pm11.327$ &  -  & -    \\
    \hline \toprule
  \end{tabular}
\end{table*}

\section{Evaluation results and ranking analysis} \label{sec:results}
Table \ref{tab:all_results} shows the MRE and SDR@2.0mm metrics for the top 10 teams.
We use HRNet \cite{wang2020deep} as the baseline since it is one of the SOTA methods for landmark detection.
It can be observed that the vanilla HRNet performance is poor in cephalometric landmark detection compared to the participants’ results. 
All 10 teams achieved better scores than HRNet in terms of MRE and SDR@2.0mm scores.
We also re-ran the ranking code by including the baseline method and found that the baseline method ranked 28th.

\begin{figure*} [htb]
     \centering
     \begin{subfigure}[b]{0.26\textwidth}
         \centering
         \includegraphics[width=\textwidth]{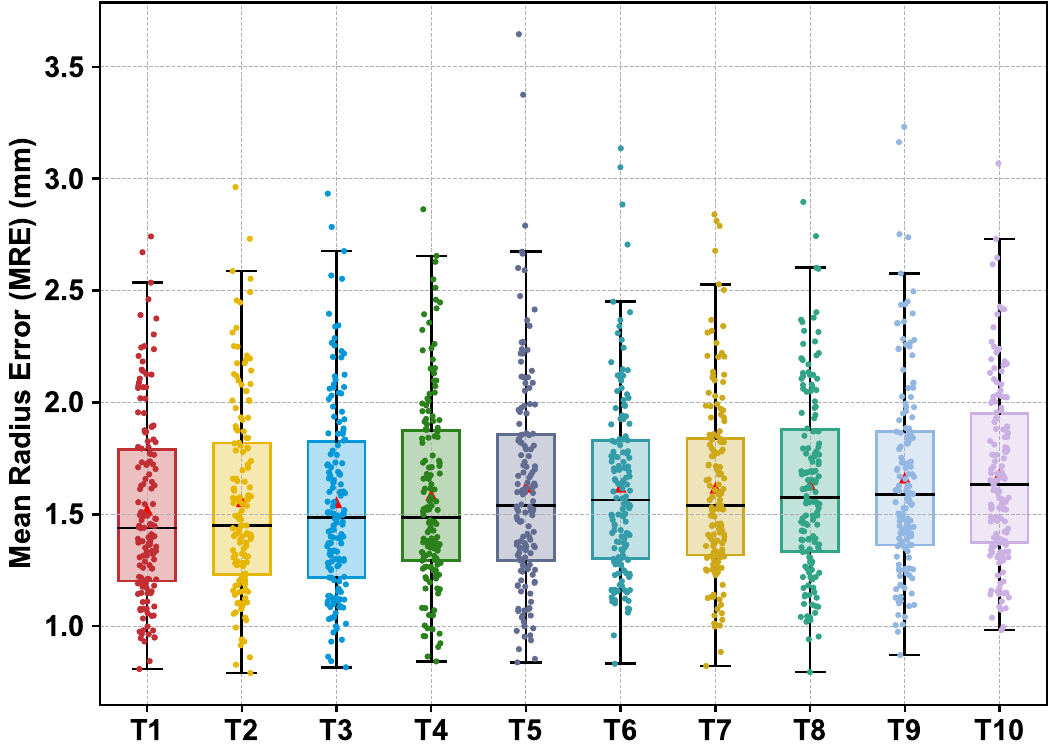}
         \caption{}
         \label{fig:test_mre_box}
     \end{subfigure}
     \hfill
     \begin{subfigure}[b]{0.201\textwidth}
         \centering
         \includegraphics[width=\textwidth]{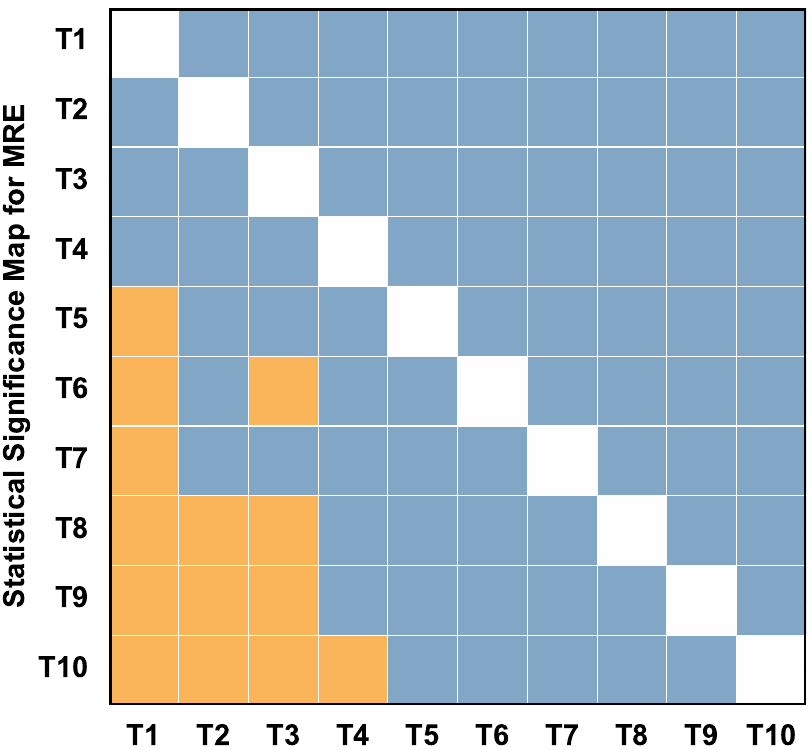}
         \caption{}
         \label{fig:test_mre_pvalue}
     \end{subfigure}
     \hfill
     \begin{subfigure}[b]{0.26\textwidth}
         \centering
         \includegraphics[width=\textwidth]{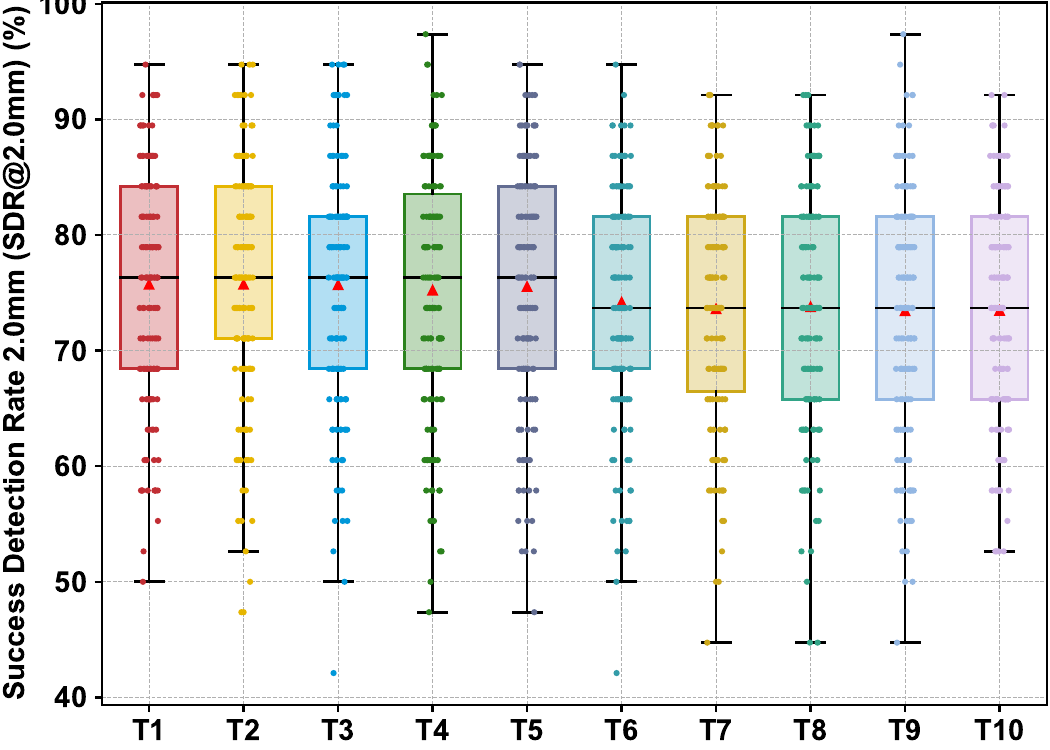}
         \caption{}
         \label{fig:test_sdr_box}
     \end{subfigure}
     \hfill
     \begin{subfigure}[b]{0.201\textwidth}
         \centering
         \includegraphics[width=\textwidth]{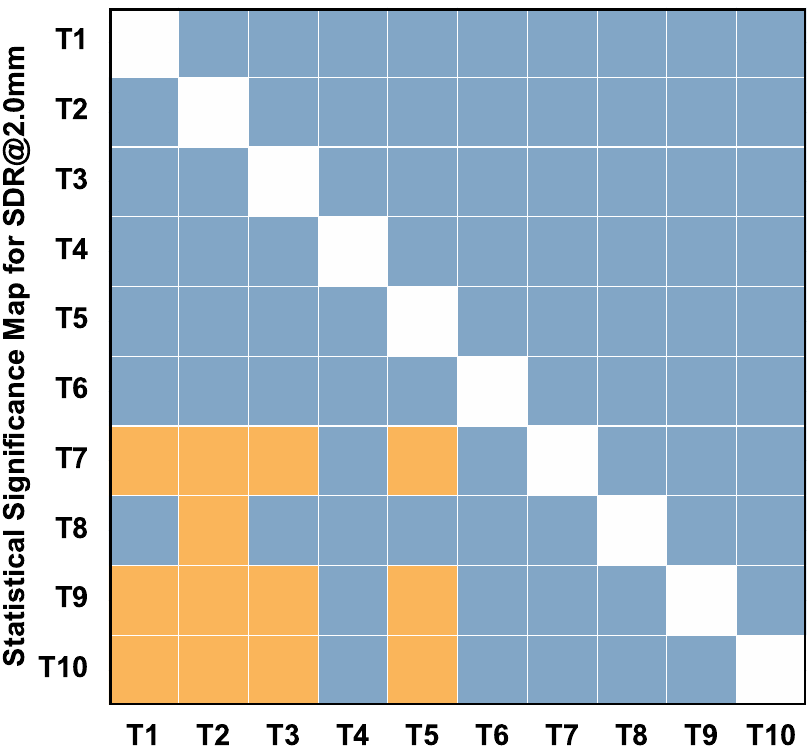}
         \caption{}
         \label{fig:test_sdr_pvalue}
     \end{subfigure}
    \caption{\textbf{Dot- and boxplot visualization (a and c) and statistical significance maps (b and d) for the MRE and SDR@2.0mm metrics of the top 10 teams.}
    (a) and (c), (b) and (d) are the results for MRE and SDR@2.0mm, respectively.
    In the statistical significance map, light yellow shading indicates that the MRE scores of the teams on the $x$-axis are significantly superior to the scores of the teams on the $y$-axis (\textit{p}-value $<$ 0.05)
    whereas light blue shading indicates they are not significantly superior.}
    \label{fig:test_result}
\end{figure*}

In the next subsections, we present the results analysis of the MRE and SDR@2.0mm metrics
by dot- and boxplots visualization and statistical significance maps, as shown in Fig. \ref{fig:test_result}.
Statistical significance maps are analyzed using the one-sided Wilcoxon signed rank test \cite{wilcoxon1992individual} at a significance level of 5\%,
which is used in many challenge results analysis \cite{luo2023efficient,ma2022fast}.
Subsequently, we perform comparative analysis for each cephalometric landmark.
A detailed description of the result analysis is provided below.

\subsection{MRE metric analysis}
As shown in Table \ref{tab:all_results}, all teams within the top 10 have MRE less than 2.0 mm.
In terms of MRE metric,
the top three teams are T1, T3, and T2, with the best result of $1.518\pm1.620$ mm.
However,
we find that a better MRE score does not always correlate with a better team ranking.
For example, T7 achieves a better MRE score than T5, but T5 has a better ranking.
Fig. \ref{fig:test_mre_box} presents the scatter diagram of MRE errors for the top 10 teams.
It can also be seen that the scores of the top teams have compact distributions with fewer outliers.

In addition, as shown in Fig. \ref{fig:test_mre_pvalue},
the MRE scores of T2, T3, and T4 have no significant difference compared to T1.
Moreover,
we could observe that there is no statistical differences in MRE metrics among closely ranked teams,
but the top teams perform better than the lower-ranked teams.
These findings highlight
that the MRE metric can serve as a direct measure for landmark detection algorithm performance.

\subsection{SDR@2.0mm metric analysis}
The SDR@2.0mm metric serves as an important measure in clinical applications.
In clinical practice,
a precision range of 2.0 mm is considered a well-accepted threshold \cite{yue2006automated,wang2016benchmark},
aligning with the evaluation criterion of our challenge.
As illustrated in Table \ref{tab:all_results},
the best-performing teams on the SDR@2.0mm metric are T7 and T2, both achieving a score of 75.719\%.
The SDR@2.0mm values among the top 10 teams show minimal variation.
As shown in Fig. \ref{fig:test_sdr_pvalue}, there is no statistical difference among the majority of teams.

Meanwhile,
we observe that although the ranking trends of participating teams are similar for SDR@2.0mm and MRE scores,
they do not always align perfectly.
For instance, 
the MRE score of T7 is lower than that of T5, while the SDR@2.0mm score of T5 is superior to T7.
These results highlight that MRE and SDR@2.0mm are complementary measures and confirm the necessity of including both in the evaluation metrics.

\begin{figure*} [hbt]
    \centering
    \includegraphics[width=1\linewidth]{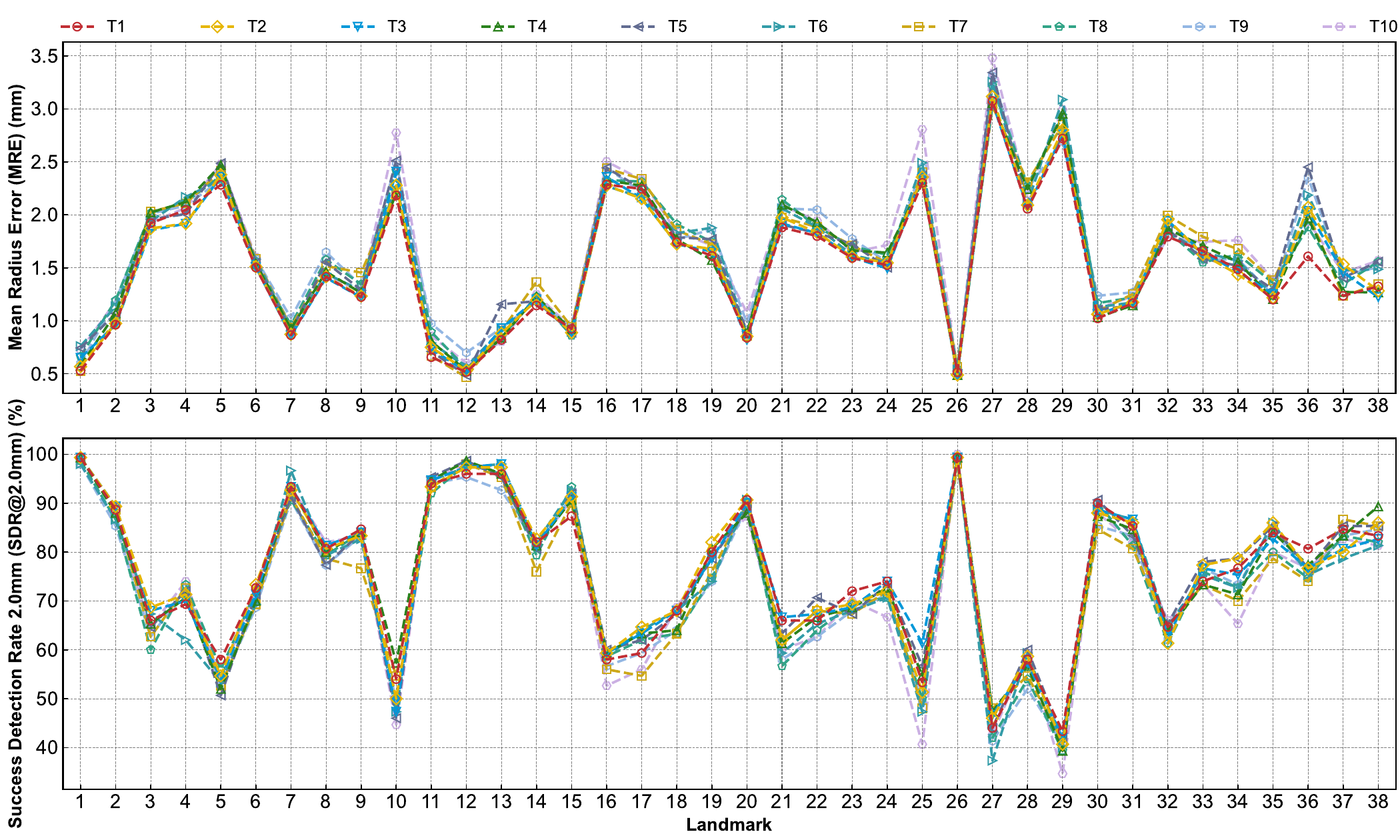}
    \caption{\textbf{Lineplot visualization for all 38 cephalometric landmark MRE and SDR@2.0mm values of the top 10 teams.}
    The upper line depicts the MRE indicator performance, while the lower line represents the SDR@2.0mm indicator performance.
    Together, these lineplots provide a comprehensive overview of the top teams' proficiency across all 38 landmarks.}
    \label{fig:all_landmark_visualization}
\end{figure*}

\subsection{Detection accuracy analysis across landmarks}
Fig. \ref{fig:all_landmark_visualization} provides a comprehensive visualization of the MRE and SDR@2.0mm values for all 38 cephalometric landmark.
The analysis of cephalometric landmarks reveals that landmark Prn (No. 26) shows the lowest error,
while landmark Basion (No. 27) poses the greatest error.
The challenge with the Basion landmark arises from the impact of overlapping skull structures, making the observation of the landmark difficult.
In addition,
the top 10 algorithms exhibit a consistent trend across landmarks.
For instance, landmarks (No. 1, 12, and 26) consistently demonstrate low MRE and high SDR@2.0mm across all algorithms,
while landmarks (No. 27 and 29) show high MRE and low SDR@2.0mm. 
Furthermore,
all algorithms consistently demonstrate better performance at corner-type landmarks compared to non-corner types.
These results suggest that the top 10 algorithms face similar challenges, lacking complementary strengths when applied in a clinical setting.

\begin{figure*} [htb]
     \centering
     \begin{subfigure}[b]{0.48\textwidth}
         \centering
         \includegraphics[width=\textwidth]{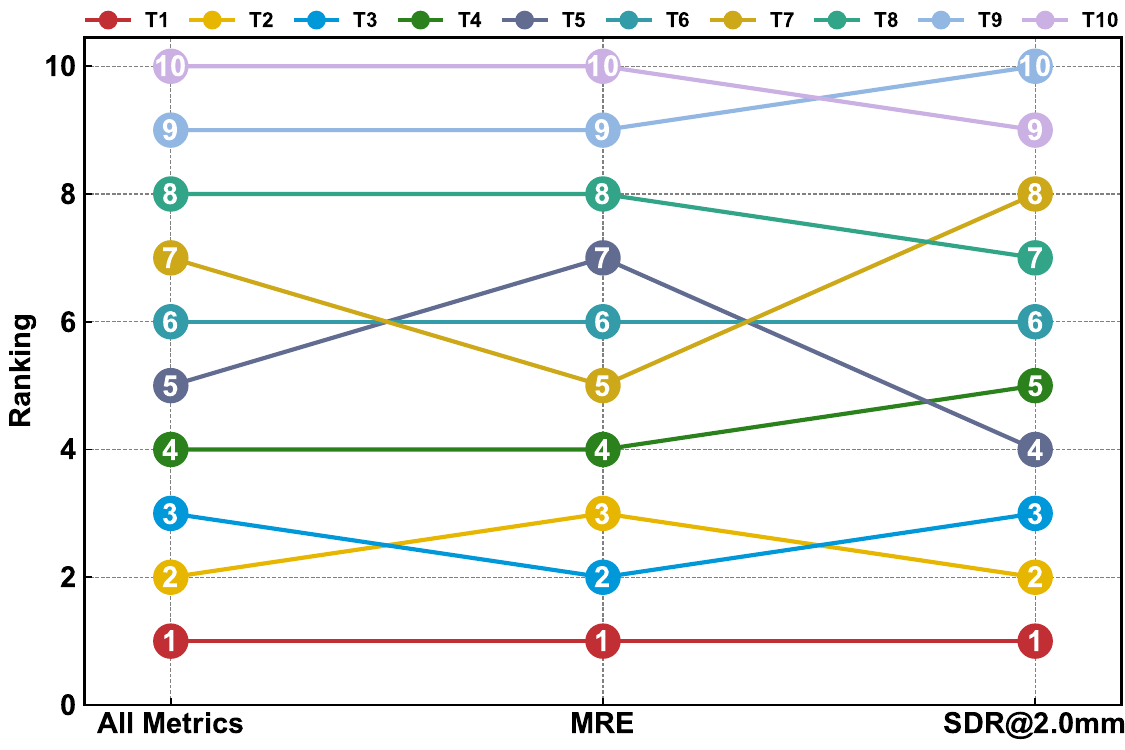}
         \caption{}
         \label{fig:ranking_plot}
     \end{subfigure}
     \hfill
     \begin{subfigure}[b]{0.48\textwidth}
         \centering
         \includegraphics[width=\textwidth]{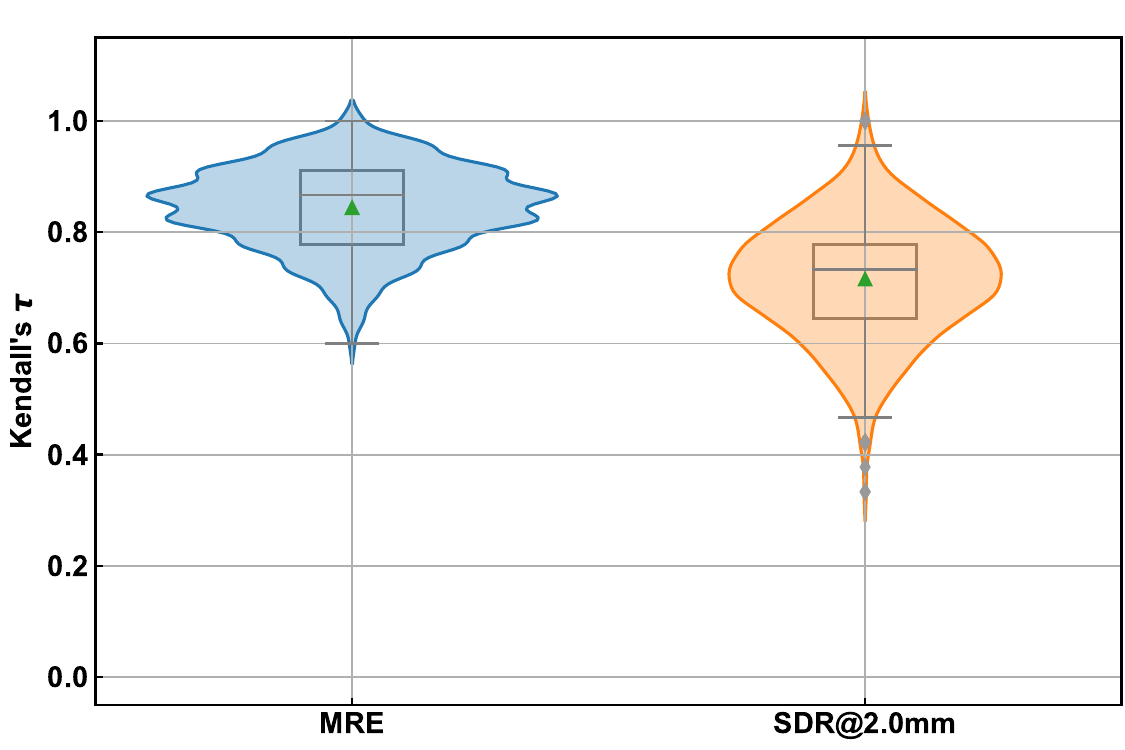}
         \caption{}
         \label{fig:ranking_kendall}
     \end{subfigure}
    \caption{\textbf{Ranking stability analysis for the top 10 teams.}
    (a) Line plots for visualizing the robustness of ranking across the top 10 teams with individual and ensemble metrics.
    According to the official ranking scheme, "All Metrics" is the ensemble of MRE and SDR@2.0mm metrics.
    (b) Violin plots for visualizing ranking stability based on bootstrapping.
    Higher values of the Kendall’s $\tau$ indicate better stability.}
    \label{fig:ranking_analysis}
\end{figure*}

\subsection{Ranking stability analysis}
To illustrate how the ranking results fluctuate with single metric,
we created a ranking chart for the top 10 teams based on the evaluation metrics, including MRE and SDR@2.0mm,
as depicted in Fig. \ref{fig:ranking_plot}.
The rankings of the top 3 teams in all metrics remain stable, demonstrating their robust performance.
However, 
lower-ranked teams exhibit significant fluctuations,
which might be attributed to different teams having different priorities. 
For example, T7 focuses on the MRE metric, while the ranking of SDR@2.0mm metric is not so favorable.
Therefore, a more comprehensive consideration of various metrics can lead to more stable ranking results. 

Besides,
we further analyzed the ranking stability of challenge ranking scheme with respect to sampling variability using a bootstrap approach \cite{kendall1938new,wiesenfarth2021methods}.
Specifically,
150 cases were randomly sampled with replacement from the testing set in 1000 repetitions.
Then, we computed the Kendall’s $\tau$ for each pair of rankings.
As described in Fig. \ref{fig:ranking_kendall},
the violin plots offer a clear of bootstrap results. 
It can be found that the values of Kendall’s $\tau$ are good,
indicating that the ranking results have high stability regardless of the testing case size.

\section{Discussion} \label{sec:discussion}
\subsection{How far are we from solving the cephalometric landmark detection problem?}

The challenge results from the top 10 teams indicate that
the top-performing methods are very close to each other with a MRE less than 2.0 mm.
In addition,
all the methods achieve a detection rate greater than 70\% using the 2.0 mm precision range, 
deemed acceptable in clinical practice.
Although these observations need to be validated on more patients,
it appears from this study that well-designed deep learning techniques can reach near-perfect detection performance.

However, conclusions are not so straightforward for the cephalometric landmark detection task.
As depicted in Fig. \ref{fig:all_landmark_visualization},
for the question: \textit{is cephalometric landmark detection a truely solved problem?},
the answers would be \textbf{NO} for A (No. 5), Go (No. 10), Pos (No. 16), PNS (No. 17), Co (No. 25), Ba (No. 27) and Bo (No. 29) landmarks detection.
Even with the same data distribution between the testing and training sets,
there are still unacceptable errors in the accurate locating of these landmarks.
This presents obstacles to the determination and implementation of the orthodontic treatment plan \cite{holdaway1956changes,fink1992duration}.
For example, the subspinale (A) landmark, the most concave point of anterior maxilla,
is used to form the ANB angle with the N and B landmarks. 
The angle serves as a crucial reference for skeletal type classification \cite{hussels1984analysis,ahmed2018diagnostic}.
Therefore, we argue that deep learning techniques for cephalometric landmark detection remains to be an unsolved problem.

In addition, as mentioned in Section \ref{sec:solutions},
we can find that, 
the majority of participating teams have concentrated their efforts on the utilization of large-resolution image inputs and the adoption of models ensemble or TTA strategies.
Despite these approaches can improve model generalization,
they come with the trade-off of high GPU usage (20+GB) and longer inference time,
which may be far from typical clinical settings.
Fortunately, as per the T2 team's solution,
incorporating deep learning modules such as separable convolution allows the model to stay lightweight without compromising performance.
This approach opens up potential avenues for overcoming these challenges and advancing the field forward.

\begin{table*} [hbt]
\centering
\caption{\textbf{Characteristics of the top 10 teams.}
Abbreviation:
a) Preprocessing: Normalization (N), Cropping (C), Resize (RS);
b) Data augmentation: Rotation (R), Flipping (F), Scaling (S), Deformation (D), Intensity transformation (IT), Mirror (M), Random cropping (RC), Random noise (RN);
c) Network design: Coarse-to-fine Framework (C2F), Feature fusion (FF), Attention Block (AB); Deep supervision (DS), Pointwise Convolution Block (PCB);
d) Inference: Test-time augmentation (TTA), Models Ensemble (ME);
e) Postprocessing: DARK \cite{zhang2020distribution} debiasing approach.}
\label{tab:characteristics}
\setlength{\tabcolsep}{2.35mm} 
\renewcommand\arraystretch{1.2} 
\begin{tabular}{cccccccccccccccccccc} 
\hline
\toprule
\multirow{2}{*}{Teams} & \multicolumn{3}{c}{Preprocessing} & \multicolumn{8}{c}{Data augmentation} & \multicolumn{5}{c}{Network design} & \multicolumn{2}{c}{Inference} & Post-processing  \\ 
\cmidrule(r){2-4}  \cmidrule(r){5-12} \cmidrule(r){13-17} \cmidrule(r){18-19} \cmidrule(r){20-20} 
                        & N & C & RS                         & R & F & S & D & M & IT & RC & RN           & C2F & FF & AB & DS & PCB          & TTA & ME                      & DARK             \\ 
\hline
T1                     
& $\checkmark$ &   & $\checkmark$                       
& $\checkmark$ &   & $\checkmark$ &   & $\checkmark$ & $\checkmark$ & $\checkmark$ &            
& $\checkmark$ &     &     &    &           
&     &  $\checkmark$                  
&                  \\ 
T2                     
& $\checkmark$ &   & $\checkmark$
& $\checkmark$ & $\checkmark$ & $\checkmark$ &   & $\checkmark$ & $\checkmark$ & $\checkmark$ &               
&     &     &     & $\checkmark$ & $\checkmark$
& $\checkmark$  &  $\checkmark$                       
& $\checkmark$                \\ 
T3                     
& $\checkmark$ & $\checkmark$ & $\checkmark$                         
&    & $\checkmark$ &   & $\checkmark$ & $\checkmark$ &   &   & $\checkmark$            
&    &     &     &    &             
&    & $\checkmark$                       
& $\checkmark$               \\ 
T4                     
& $\checkmark$ &   & $\checkmark$ 
& $\checkmark$ & $\checkmark$ & $\checkmark$ &   &   & $\checkmark$ & $\checkmark$ & $\checkmark$            
&     &  $\checkmark$ &     &    &            
& $\checkmark$ &                         
&                  \\ 
T5                     
& $\checkmark$ &   & $\checkmark$                         
&    & $\checkmark$ &   &   &   &   &   &              
&    & $\checkmark$ &     &    &             
&    & $\checkmark$                       
&                  \\ 
T6                     
& $\checkmark$ & $\checkmark$ & $\checkmark$                         
& $\checkmark$ &   &   &   &   & $\checkmark$ & $\checkmark$ &              
&     &  $\checkmark$  &     &    &              
& $\checkmark$ & $\checkmark$                        
&                  \\ 
T7                     
& $\checkmark$ & $\checkmark$ & $\checkmark$                         
& $\checkmark$ & $\checkmark$ &   &   &   &   &   & $\checkmark$             
& $\checkmark$ &     &     &    &             
&     &                         
&                  \\ 
T8                     
& $\checkmark$ &   & $\checkmark$                        
&    &   &   & $\checkmark$ &   & $\checkmark$ &   &              
&     &     &     &    &             
&     & $\checkmark$                       
&                  \\ 
T9                     
& $\checkmark$ & $\checkmark$ & $\checkmark$                         
& $\checkmark$ & $\checkmark$ &   &   &   &   &   &              
&     & $\checkmark$ & $\checkmark$ &    &             
&     &                         
&                  \\ 
T10                    
& $\checkmark$ & $\checkmark$ & $\checkmark$                         
&    & $\checkmark$ &   &   & $\checkmark$ &   &   &              
&     &     &     &    &              
&     & $\checkmark$                        
&                  \\
\hline
\toprule
\end{tabular}
\end{table*}

\subsection{What strategies can help you stand out?} \label{sec:what_help}
To better understand the strategies behind the success of top-performing teams,
we present a comprehensive summary of the characteristics observed among the top 10 teams in Table \ref{tab:characteristics}. 
Then, we conduct an in-depth analysis of common approaches as outlined below.

\begin{itemize}
\item \textbf{Data Preprocessing:} During data preprocessing, most top teams normalized (N) intensity values into $[0, 1]$, which reduced the intensity variances among different cases and centers.
Since the high resolution of original lateral X-ray images, exceeding $2000 \times 2000$ without uniform sizes,
it can hardly be fed directly into the network because of the huge memory consumption.
To address this,
all the top 10 teams employed a resize (RS) operation to standardize image sizes.
Besides, some teams (e.g. T3, T6 and T7) further conducted a crop (C) operation to eliminate interference from redundant areas prior to resizing.
All these strategies could help increase landmark detection accuracy and efficiency.

\item \textbf{Data Augmentation:} Extensive data augmentation (e.g.
rotation (R), flipping (F), scaling (S), deformation (D), intensity transformation(IT), mirror (M), random cropping (RC), random noise (RN), etc.) were also used by most top teams (Table  \ref{tab:characteristics} Data Augmentation),
which were effective ways to improve detection accuracy on the testing cases with different skull appearances and shapes.
For example,
in contrast to T10, which did not perform much data augmentation, 
T3, with extensive data augmentation, enhanced the generalization of the CL-Detection dataset,
leading to better performance.

\item \textbf{Model Paradigm:}
All the top 10 teams adopt heatmap-based method to detect landmarks.
The success of this paradigm lies in its ability to effectively capture landmark information in images using heatmaps,
resulting in more precise and reliable landmark detection \cite{zhu2021you,zhou2021learn}.
Furthermore, through heatmap visualization, participants can gain a more intuitive understanding of the model's outputs, leading to optimization and improvement in the accuracy and efficiency of landmark detection.
Therefore, the successful application of this heatmap-based model paradigm in landmark detection demonstrates its unique advantages in enhancing algorithm performance and driving research progress.

\item \textbf{Network Backbone:}
Among the top 10 teams,
four teams constructed their networks based on encoder–decoder architecture U-Net \cite{ronneberger2015u} or its improved variant.
For instance,
T1 selected EfficientNet \cite{tan2019efficientnet} as the U-Net encoder to enhance the image extraction capability to achieve a SOTA performance.
On the other hand, the remaining six teams opted for HRNet \cite{wang2020deep} as backbone network, designed to handle high-resolution input (Table \ref{tab:top10} Network).
These observations are also consistent with the current SOTA methods for landmark detection \cite{zhu2021you}.

\item \textbf{Network Design:}
Two teams, T1 and T7, adopt coarse-to-fine (C2F) framework (Table. \ref{tab:characteristics} Network design).
It is a popular choice in many medical landmark detection tasks \cite{zhang2023craniomaxillofacial,cheng2023prior}.
In addition,
there were four teams among the top 10 teams that utilized feature fusion (FF) modules to fuse different scale features,
contributing to improved detection performance.
Furthermore,
deep supervision (DS), attention block (AB) and pointwise convolution block (PCB)
were also explored by the top teams.
For example, T2 integrated DS and PCB techniques into HRNet model to enhance the landmark detection performance beyond the capabilities of the baseline model.

\item \textbf{Inference:}
During the inference phase,
the frequently employed method in the CL-Detection2023 challenge was model ensemble (ME), as shown in Table. \ref{tab:characteristics}.
Among the top 10 teams, seven teams leveraged ME method to improve the generalization of the model.
Besides,
T2, T4 and T6 applied TTA strategy.
For example, left-right flipped TTA was used by T4 to address the issue of landmark deviation randomness.

\item \textbf{Post-processing:}
Heatmap-based landmark detection methods often use softmax method to decode the heatmap and obtain the coordinates of the landmarks.
However, this introduces bias into the estimation.
In response to this issue,
T2 and T3 implemented DARK debiasing scheme \cite{zhang2020distribution} to decode landmark coordinates.
This adjustment led to a noteworthy improvement of approximately 1\% in SDR@2.0mm metric \cite{wu2023revisiting}.

\end{itemize}

\begin{table*} [!t]
\centering
\caption{
\textbf{Quantitative evaluation results of MRE metric for the detection of three landmarks with large errors in three medical centers using the CL-Detection2023 Challenge SOTA method.}
The ANOVA statistical method \cite{student1908probable} was applied to compare the means across different centers ($p$-value $<$ 0.05).}
\label{tab:center_test}
\renewcommand\arraystretch{1.2} 
\setlength{\tabcolsep}{6mm}  
\begin{tabular}{lcccccc}
\hline
\toprule
Landmarks          &  Overall          &  Center1          &  Center2          & Center3          & ANOVA test $p$-value  \\ \hline
No. 16 (Pos)  &  $2.285\pm2.120$  &  $2.414\pm2.420$  &  $2.079\pm1.021$  & $1.915\pm1.230$  & 0.520       \\ 
No. 29 (Bo)   &  $2.720\pm1.809$  &  $2.786\pm1.713$  &  $2.078\pm1.665$  & $3.024\pm2.170$  & 0.169        \\ 
No. 27 (Ba)   &  $3.076\pm2.363$  &  $3.457\pm2.471$  &  $1.770\pm1.594$  & $2.621\pm1.928$  & \textbf{0.005}       \\ \hline
\toprule
\end{tabular}
\end{table*}

\subsection{Where do methods fail?}
In the light of the results reported so far,
it seems that top deep learning detection methods fall within the range of human expectations based on MRE scores.
Nonetheless, as depicted in Fig. \ref{fig:all_landmark_visualization},
there remains a difference of 2 to 4 mm from expert performance in specific landmarks.
This prompts the question: where do these methods fail?

As evident in Table \ref{tab:data} and Fig. \ref{fig:challenge_c},
there are non-negligible imbalances and distribution variations in data obtained from different centers.
Therefore,
one hypothesis can be that algorithms trained on imbalanced datasets might struggle to generalize effectively to test data from particular centers, thereby yielding substantial errors in certain landmarks.
To verify this assumption,
we broke down the MRE metric of the top-performing T1 team for each center.
As presented in Table. \ref{tab:center_test},
we focus on the three most challenging landmarks (Pos, Bo and Ba),
and adopt the ANOVA statistical method \cite{student1908probable} to compare their means.
As one can see,
there is still an issue with the data distribution of some landmarks (e.g. Ba) in different medical centers, causing the method to systematically fail.
However,
not all landmarks are affected by the data distribution, leading to differences in results across centers.
One reason for this could be explained by the fact that images from different centers are included in the training set thus allowing deep networks to learn center-specific representations.

\begin{table} [!t]
\centering
\caption{\textbf{Comparison of quantitative results for soft tissue-related landmarks and bone-related landmarks.}
T-test ($p$-value $<$ 0.05) was used for significance testing.}
\label{tab:soft_vs_bone}
\renewcommand\arraystretch{1.2} 
\setlength{\tabcolsep}{4.8mm}  
\begin{tabular}{lcc}
\hline
\toprule
Landmark type       & MRE (mm)         & SDR@2.0mm (\%)         \\ \hline
Soft tissue-related  & $1.324\pm1.564$  & $80.769\pm14.988$       \\
Bone-related         & $1.618\pm1.642$  & $73.093\pm10.841$       \\\hline
$p$-value            & $>0.999$         & $>0.999$                \\ \hline
\toprule
\end{tabular}
\end{table}

\begin{figure*} [!t]
    \centering
    \includegraphics[width=\linewidth]{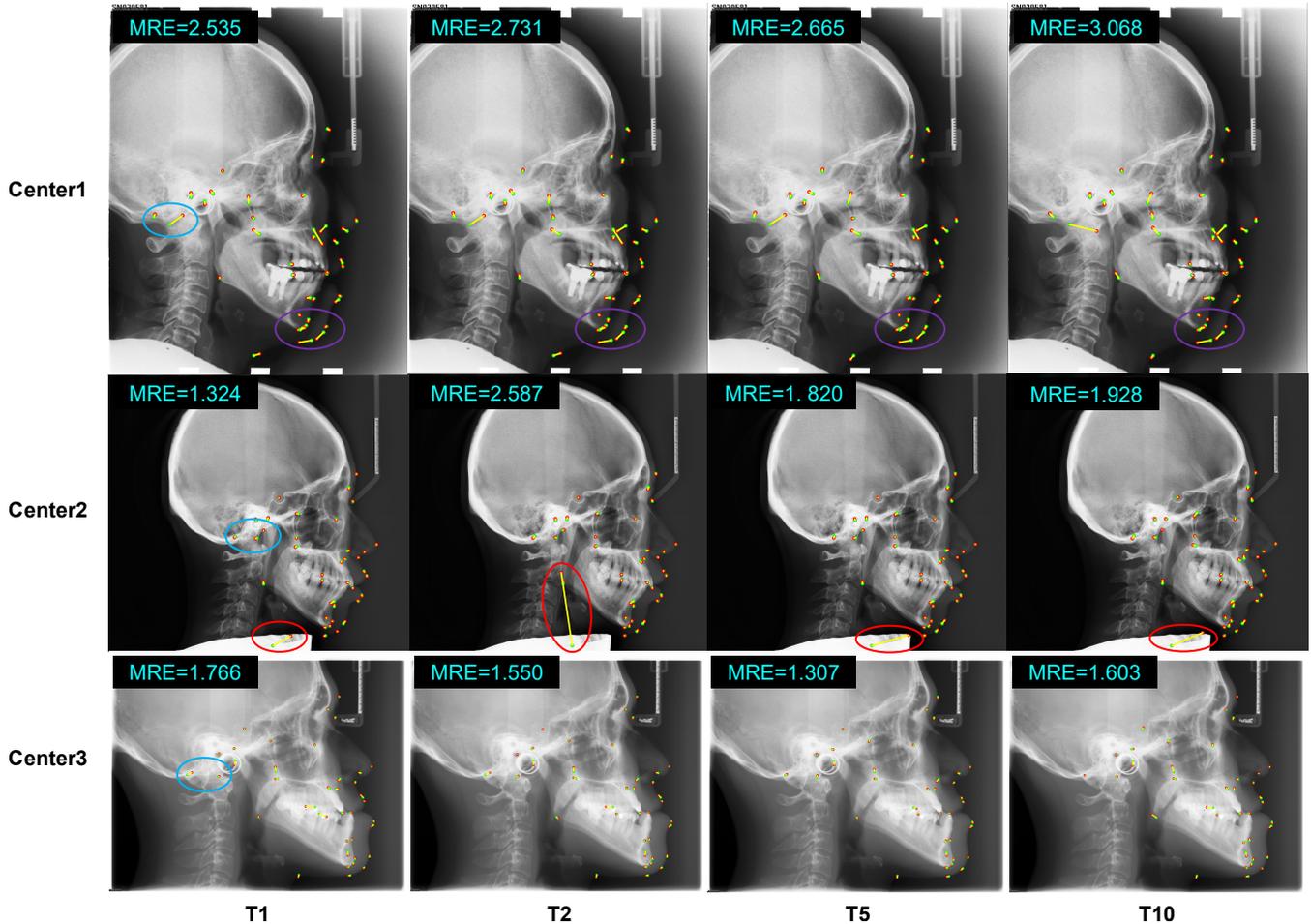}
    \caption{\textbf{Landmark visualization results in different medical centers of the different four teams (T1, T2, T5 and T10).}
    The green point and red point denote the ground truth and prediction landmarks, respectively.
    The yellow line is the line between ground truth landmark and prediction landmark.}
    \label{fig:visualize_cases}
\end{figure*}

Another hypothesis suggests that soft tissue-related landmarks are more difficult to localize than bone-related landmarks due to low contrast in nearby areas \cite{mostafa2009soft,giannopoulou2020orthodontic}.
However, upon a thorough analysis of the detection results,
we observed no significant differences between the outcomes for soft tissue-related and bone-related landmarks,
as demonstrated in Table \ref{tab:soft_vs_bone}. 
Furthermore, it is unexpected that
soft tissue-related landmarks are easier to detect than bone-related landmarks as the scores relative to this group get the larger MRE and lower SDR@2.0mm.
As seen from the comparison of landmarks in the purple and blue areas in Fig. \ref{fig:visualize_cases},
soft tissue-related landmarks are influenced by low contrast, but bone-related landmarks face challenges due to the overlapping nature of skull bones.
This complexity makes it difficult to determine definitively which type of landmark is more easily detectable.

Another hypothesis generally accepted in the machine learning community is that
heatmap-based landmark detection methods overlook the topological structure between landmarks \cite{li2020structured}.
As illustrated in the first row of Fig. \ref{fig:visualize_cases},
the relative positions of some predicted landmarks do not conform to the expected anatomical relationships.
This may be because the top 10 algorithms are all heatmap-based solutions.
During heatmap decoding, each landmark's heatmap is processed individually without considering the topological connections between landmarks.
This oversight leads to inherent errors in the final post-processing stage.
Despite the availability of post-processing techniques like DARK \cite{zhang2020distribution} that aim to mitigate these issues, such discrepancies remain unavoidable.

In addition,
the comparison of the landmarks in the red circle of Fig. \ref{fig:visualize_cases} reveals that different ranking algorithms fail to execute when landmarks are obstructed, leading to significant location errors.
This issue may stem from the CNN-based algorithms excelling at capturing local information but failing in perceiving occluded landmarks \cite{viriyasaranon2023anatomical}.
Moreover, the intrinsic shape constraints inherent in heatmap-based methods \cite{li2020structured} are absent, resulting in the predicted landmarks appearing at inappropriate locations.
However, it is worth pointing out that 
the algorithm designed by T1 has the smallest error in this case compared to other teams.
Therefore, we believe that 
the detection of occluded landmark points would be solved through more sophisticated design algorithms, such as graph convolution that are good at handling topological structures or a Transformer model that are long-range aware.

\subsection{For the need of new metrics}
As the saying goes, success and failure are often two sides of the same coin.
The results presented in Table \ref{tab:all_results} indicate that top deep learning methods are close to the expert level.
However, as discussed in Section \ref{sec:what_help},
most top teams adopt ME methods or TTA strategies to improve model performance.
These techniques produce large model sizes and consume computational resources, limiting the application scenarios of the developed deep learning models.
For improved AI applications in science,
we advocate for the inclusion of GPU usage and runtime supervision in future endeavors related to landmarks localization.
This may involve considering new evaluation metrics such as program running time and the area under GPU memory-time curve, as proposed by the FLARE challenge \cite{ma2022fast}.
This approach may guide the algorithm to strike a balance between effectiveness and efficiency.

\subsection{Limitations and future work}
In our CL-Detection2023 challenge dataset, each case has 38 landmarks annotation but the testing cases do not have different annotations from multiple raters.
However, the data annotations were verified by a senior doctor with over 20 years of experience.
This review process significantly minimize the risk of personal bias in the annotations.
Moreover,
our ranking scheme does not include runtime-related and GPU-related measures,
which allow participants to obtain advantages by using ME or TTA method.
Nevertheless, we have made the evaluation measures and code, and ranking scheme publicly available on the challenge website.
Thus, it is fair for participants to choose these inference methods.

In addition,
our challenge data only contains a small amount ($<10\%$) of children's X-ray data, and it is uncertain whether the current top-performing algorithm is still robust, especially for tooth-related landmarks. 
This is because a child's mouth is always affected by permanent teeth, which can make landmarks difficult to identify.
Furthermore, the findings presented in Table \ref{tab:center_test} demonstrate the substantial influence that varying distribution data from different medical centers can have on specific landmarks within the algorithm. 
Considering that the current data scale is not large,
but it is already the largest and most diverse open source cephalometric landmark dataset, 
we mixed three medical centers' data to divide the training, validation and testing sets.
Therefore, in future research,
we will collect a child-focused cephalometric data set to serve as an independent test set to thoroughly explore the domain-adaption issues associated with "different medical centers" and "adult-child" variations.

\section{Conclusion} \label{sec:conclusion}
In conclusion,
we have curated a large-scale and diverse cephalometric lateral X-ray dataset and organized an international challenge to validate and compare the performance of deep learning landmark detection algorithms.
The quantitative results show that 
the winning method in our challenge achieved a distance error of 1.518 mm and a successful detection rate of 75.719\% for 2.0 mm precision range.
This result also demonstrates that state-of-the-art deep learning methods can successfully get highly accuracy cephalometric landmark detection results.
However, the top-performing methods are still fails at the occlusion and low-contrast landmarks,
especially when considering the distance error.
Looking ahead, we expect to be able to develop more sophisticated methods and obtain higher successful detection rates based on the data we constructed.
We also hope that this work will provide valuable insights for future algorithm development.

\section*{Declaration of Competing Interest}
The authors declare that they have no known competing financial interests or personal relationships that could have appeared to influence the work reported in this paper.

\begin{figure*} [!t]
    \centering
    \includegraphics[width=\linewidth]{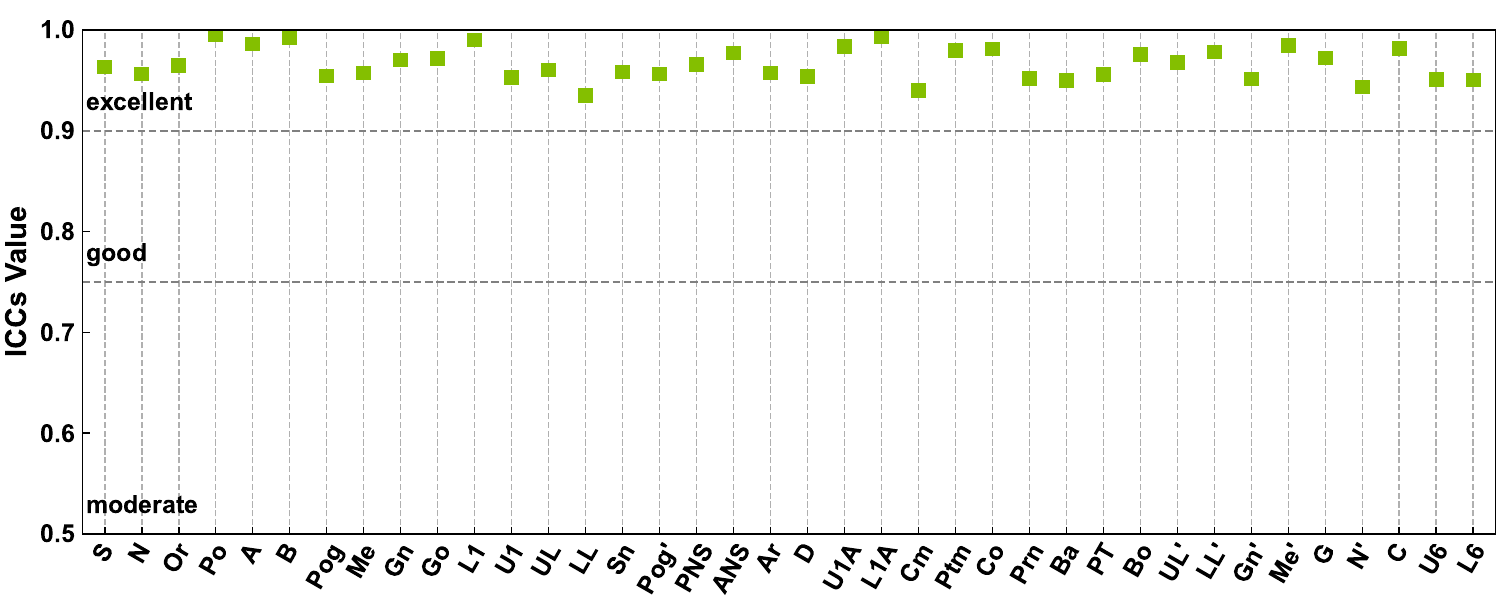}
    \caption{Intraclass Correlation Coefficients (ICC) between the two senior doctors. Each point on the figure represents the ICC of a particular landmark. The green point indicates that the annotated landmarks have excellent agreement.}
    \label{fig:senior_icc}
\end{figure*}

\section*{Appendix A. Inter-observer variability evaluation}
To verify the variability between the two doctors involved in the landmark annotation,
we have calculated the inter-observer variability of these two senior doctors on a subset of 50 cases to assess human performance.
Two dental specialists annotated 38 cephalometric landmarks for each digital lateral cephalometric X-rays. Interclass Correlation Coefficients (ICCs) \cite{bartko1966intraclass,shrout1979intraclass} were calculated to assess inter-observer variation.
According to the general guidelines for ICC measure \cite{KOO2016155},
an ICC $>$ 0.90 indicates excellent agreement, an ICC of 0.75 - 0.90 reflects good agreement, and an ICC $<$ 0.75 represents poor to moderate reliability.
Current findings suggest that ICCs are excellent for senior doctor, and ICCs over 0.90 for all landmarks (Fig. \ref{fig:senior_icc}).

\section*{References}
\bibliographystyle{IEEEtran}
\bibliography{refs}

\end{document}